\pdfoutput=1 

\documentclass{article} %
\usepackage[table]{xcolor}

\usepackage{iclr2025_conference,times}
\iclrfinalcopy %

\usepackage{amsmath,amsfonts,bm}

\def\eqref#1{equation~\ref{#1}}

\def\1{\bm{1}}

\def\ermS{{\textnormal{S}}}

\DeclareMathAlphabet{\mathsfit}{\encodingdefault}{\sfdefault}{m}{sl}
\SetMathAlphabet{\mathsfit}{bold}{\encodingdefault}{\sfdefault}{bx}{n}

\usepackage{hyperref}
\usepackage{url}

\usepackage{graphicx}
\usepackage{caption}
\usepackage{amsmath}
\usepackage{amssymb}
\usepackage{booktabs}
\usepackage{makecell}
\usepackage[export]{adjustbox}
\usepackage[normalem]{ulem}
\usepackage{textpos}  %
\usepackage{wrapfig}

\usepackage{fontawesome}

\usepackage{adjustbox}

\usepackage{tabularx}
\usepackage{subcaption}

\usepackage{tabulary,multirow,overpic,subfloat}
\usepackage{xspace}

\newcommand{\app}{\raise.17ex\hbox{$\scriptstyle\sim$}}

\newcolumntype{x}[1]{>{\centering\arraybackslash}p{#1pt}}
\newcolumntype{y}[1]{>{\raggedright\arraybackslash}p{#1pt}}

\newlength\savewidth\newcommand\shline{\noalign{\global\savewidth\arrayrulewidth
  \global\arrayrulewidth 1pt}\hline\noalign{\global\arrayrulewidth\savewidth}}
\newcommand{\tablestyle}[2]{\setlength{\tabcolsep}{#1}\renewcommand{\arraystretch}{#2}\centering\footnotesize}
\makeatletter\renewcommand\paragraph{\@startsection{paragraph}{4}{\z@}
  {.5em \@plus1ex \@minus.2ex}{-.5em}{\normalfont\normalsize\bfseries}}\makeatother

\newcommand\blfootnote[1]{\begingroup\renewcommand\thefootnote{}\footnote{#1}\addtocounter{footnote}{-1}\endgroup}

\newcommand{\modelname}{MM-Ego\xspace}
\newcommand{\benchname}{EgoMemoria\xspace}

\title{\modelname: Towards Building Egocentric \\ Multimodal LLMs for Video QA}

\author{$^\text{\faApple}$Hanrong Ye\textsuperscript{1}\footnotemark[2]\>\,, Haotian Zhang\textsuperscript{2}\footnotemark[2]\>\,, Erik Daxberger\textsuperscript{2}, Lin Chen\textsuperscript{2}, $^\text{\faApple}$Zongyu Lin\textsuperscript{3}, Yanghao Li\textsuperscript{2}, \\
\textbf{Bowen Zhang\textsuperscript{2}, Haoxuan You\textsuperscript{2}, Dan Xu\textsuperscript{1},  Zhe Gan\textsuperscript{2}\footnotemark[3], Jiasen Lu\textsuperscript{2}\footnotemark[3], Yinfei Yang\textsuperscript{2}\footnotemark[3]}  \\ 
\textsuperscript{1}CSE, HKUST \quad \textsuperscript{2}Apple \quad \textsuperscript{3}UCLA \\
\small{\texttt{hanrong.ye@connect.ust.hk}, \texttt{\{haotian.zhang2,yinfeiy\}@apple.com}} \\
}

\begin{document}

\maketitle

\begin{figure*}[h]
    \centering
\includegraphics[width=1.0\linewidth]{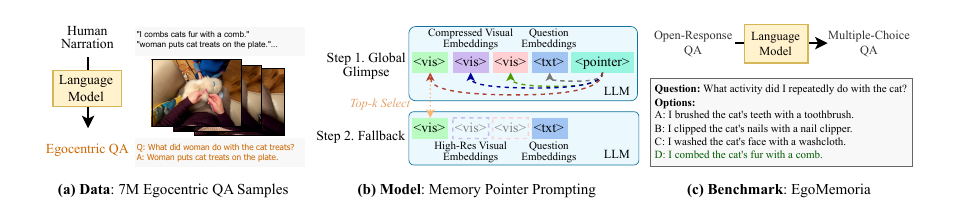}
\caption{We introduce a foundation model for egocentric video understanding, contributing from three key perspectives: (a) 7 million egocentric QA samples generated from human narrations via a data engine, (b) a multimodal language model designed for egocentric video comprehension, and (c) the curation of a challenging egocentric video understanding benchmark. 
}
\label{fig:overall}
\end{figure*}

\begin{abstract}
This research aims to comprehensively explore building a multimodal foundation model for egocentric video understanding.
To achieve this goal, we work on three fronts. 
First, as there is a lack of QA data for egocentric video understanding, we automatically generate 7M high-quality QA samples for egocentric videos ranging from 30 seconds to one hour long in Ego4D~\citep{grauman2022ego4d} based on human-annotated data.
This is one of the largest egocentric QA datasets.
Second, we contribute a challenging egocentric QA benchmark with 629 videos and 7,026 questions to evaluate the models' ability in recognizing and memorizing visual details across videos of varying lengths. We introduce a new de-biasing evaluation method to help mitigate the unavoidable language bias present in the models being evaluated.
Third, we propose a specialized multimodal architecture featuring a novel ``Memory Pointer Prompting" mechanism. This design includes a \textit{global glimpse} step to gain an overarching understanding of the entire video and identify key visual information, followed by a \textit{fallback} step that utilizes the key visual information to generate responses. This enables the model to more effectively comprehend extended video content.
With the data, benchmark, and model, we build MM-Ego, an egocentric multimodal LLM that shows powerful performance on egocentric video understanding.
\blfootnote{$^\text{\faApple}$Work done during an internship at Apple. 
$^\dagger$First Authors.
$^\ddagger$Senior Authors.
}
\end{abstract}

\section{Introduction}
\label{sec:intro}
Study on egocentric videos explores how machines can see and understand the world from a first-person, self-centered perspective.
Egocentric videos differ significantly from static-camera videos, such as movies or animations, both in terms of content and viewpoint. 
The content of egocentric videos primarily revolves around human daily activities. 
These videos typically share a perspective similar to human vision, where the camera and viewpoint frequently move. As a result of these characteristics, egocentric videos exhibit a distinct data distribution compared to static-camera videos, which has motivated a new area of research. 
In recent years, research interest in egocentric intelligence has been on the rise~\citep{sigurdsson2018charadesego,Damen2018EPICKITCHENS,grauman2022ego4d,mangalam2023egoschema,plizzari2024outlook}. 
This growing interest is driven by the rapid advancements in AR/VR headsets and robotics, where cameras capture long-form egocentric videos in a manner akin to human vision. 
Research on egocentric videos will allow these devices to understand their surroundings and human intentions, fostering more advanced machine intelligence and improving the human-machine interaction experience, with immeasurable research and application potential.

However, research on understanding egocentric videos remains in its early stages, with previous research primarily centered on specialized tasks such as story summarization~\citep{lee2012discovering}, hand-object relationship understanding~\citep{cai2016understanding}, action classification~\citep{cartas2017recognizing,li2021ego}, and temporal or spatial grounding~\citep{grauman2022ego4d}. In contrast, works focusing on developing a more general egocentric video understanding model capable of complex understanding remain rare.
Despite that video multimodal large language models (MLLMs) demonstrate strong video understanding and reasoning ability~\citep{damonlpsg2023videollama,wang2024internvideo2,lin2023vila,zhang2024llavanextvideo}, most of these works are unsuitable for egocentric video understanding from data, benchmark, and model design perspectives. 

$(a)$ From a data standpoint, although many MLLMs use some egocentric videos from ActivityNet~\citep{yu2019activitynet}, Ego4D~\citep{grauman2022ego4d}, and Charades~\citep{sigurdsson2018charadesego} in their training,
they have not been trained on \textit{large-scale} egocentric video datasets, which inherently restricts their ability to comprehend lengthy first-person videos and accurately extract visual details. While Ego4D~\citep{grauman2022ego4d} offers valuable human-annotated videos and labels for certain egocentric video understanding tasks, particularly episodic memory (which assesses a model's ability to retain visual details in such videos), its annotations are not structured for generating language responses, making them unsuitable for training MLLMs. Therefore, a large-scale egocentric video QA corpus is still needed. 
$(b)$ In terms of benchmarking, exisiting video QA benchmarks either focus on shorter videos -- such as EgoSchema~\citep{mangalam2023egoschema} and QaEgo4D, which evaluate using around 3-minute and 8-minute videos, respectively -- or concentrate on Internet video content~(e.g., Video-MME~\citep{fu2024videomme}). This creates a notable gap in egocentric video understanding benchmarks that encompass videos ranging from seconds to an hour in length.
$(c)$ From a model design perspective, 
previous video MLLMs have primarily addressed long videos in two ways.  
The first approach involves uniformly sampling a limited number of video frames as visual input, as seen in \citet{li2024llavaov,lin2023vila}.
Despite its simplicity, this approach achieves better performance among open-source models on public video benchmarks~\citep{fu2024videomme}, largely because its design ensures high training efficiency and good scaling properties.
The second approach involves feeding a large volume of visual tokens into the transformer backbone and employing engineering techniques, such as tensor parallelism and sequence parallelism~\citep{xue2024longvila,zhang2024longva}, to facilitate training with millions of visual tokens in context.
However, these long-context transformers suffer from slow training speeds and small overall batch sizes, which hinder performance improvements given the constraints of computational resources and training time.
Intuitively, even humans cannot remember every detail of an hour-long video. We believe a more effective approach is to understand the video progressively: first get an overview of the entire video, then focus on specific details with particular questions in mind.

Building on the observations mentioned above, we introduce \modelname, an egocentric MLLM designed to process and understand long egocentric videos. Our contributions are threefold:

($i$) \textbf{Data.}
To scale training data for MLLMs with egocentric understanding ability, we develop an efficient data engine, using a ``narration to egocentric QA'' strategy, to automatically synthesize a large-scale egocentric QA dataset based on video narration data. 
Notably, rather than relying on existing vision-language models (VLMs) as labelers, we generate egocentric QAs based on the human-annotated fine-grained video clip narrations. This approach, conceptually related to~\citep{di2024groundvqa,li2024llamavid}, ensures that our data quality is not constrained by the limitations of existing labeling VLMs.
In this way, we create one of the first large-scale egocentric QA datasets, consisting of over 7 million egocentric QA samples that span video lengths from seconds to over an hour. This dataset enables the training of models to recognize and retain visual details from egocentric videos.

$(ii)$ \textbf{Benchmark.} To evaluate the MLLMs' performance in understanding and memorizing visual details from egocentric videos, we propose the \benchname benchmark. This challenging benchmark includes 7,026 multiple-choice questions for 629 egocentric videos ranging from 30 seconds to 1 hour. 
In the experiments on \benchname, we further investigate the impact of inevitable language biases across different models during evaluation and introduce a debiased metric to more accurately assess the models' true egocentric understanding capabilities.

($iii$) \textbf{Model.} 
For our \modelname model, we develop a progressive approach to handle egocentric videos by introducing a Memory Pointer Prompting method.
It consists of two steps: ``global glimpse'' and ``fallback". 
In the \textit{global glimpse} step, we extract compressed frame-level visual embeddings from the entire video to get a global understanding. Then, we employ a memory pointer embedding, designed to examine all compressed frame-level visual embeddings along with the question embeddings, to aid in identifying key visual embeddings in a question-aware manner. 
In the following \textit{fallback} step, the selected key visual embeddings, in a higher-resolution form, are then used as final input to the LLM for processing and generation.
This approach allows us to achieve a global understanding of the entire video while also identifying and utilizing key visual information to answer questions related to visual details.

\section{Method}

\begin{figure}[t]
    \centering
    \vspace{-40pt}
    \includegraphics[width=\textwidth,clip]
    {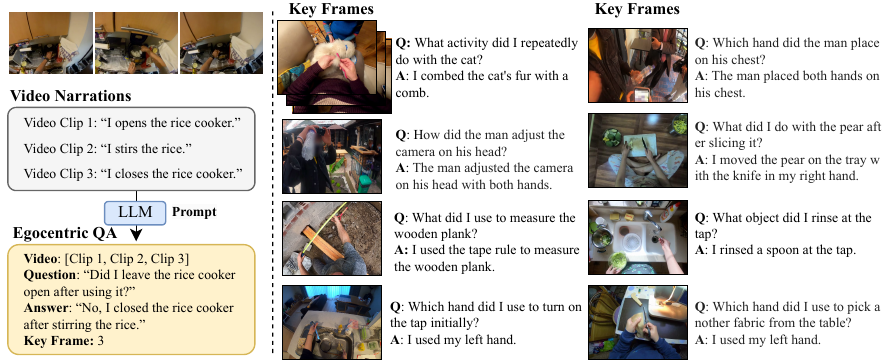}
    \vspace{-15pt}
    \caption{``Narration to Egocentric QA'' data engine. Given a sequence of human-annotated video narrations, we instruct a language model (GPT-4o) to generate egocentric understanding-related questions and answers, along with identifying the key frames necessary to answer those questions.
    }
    \label{fig:ego_mm_dataset}
    \vspace{-10pt}
\end{figure}

\subsection{``Narration to Egocentric QA'' Data Engine}
\label{sec:egomm_dataset}

As outlined in Section~\ref{sec:intro}, high-quality egocentric QA pairs are lacking for training an MLLM with egocentric video understanding ability. To address this gap, we develop an innovative ``narration to egocentric QA" data engine that automatically generates episodic memory-related QA samples based on human-annotated video clip narrations from the Ego4D dataset~\citep{grauman2022ego4d} without the need for additional manual annotations. 

\begin{wrapfigure}{r}{0.5\textwidth}
\vspace{-10 pt}
  \begin{center}
     \includegraphics[width=0.5\textwidth]{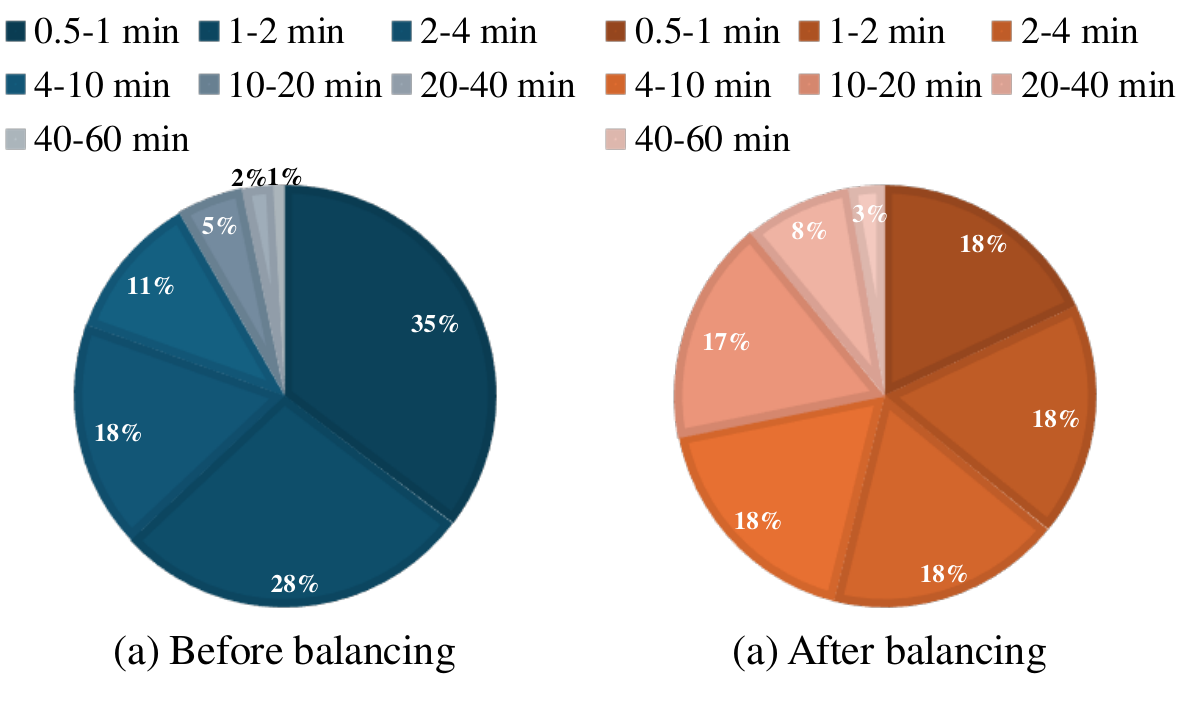}
     \vspace{-18pt}
  \end{center}
  \vspace{-0.2cm}
   \caption{Video length distribution in our egocentric QA dataset.}
  \label{fig:mm_memory_stat}
  \vspace{-0.4cm}
\end{wrapfigure}

Our approach leverages over 3,000 hours of privacy-protected, de-identified egocentric videos accompanied by more than 3 million high-quality, human-created narrations. These fine-grained language descriptions provide a rich resource for generating QA pairs.

The workflow of the data engine is illustrated in Figure~\ref{fig:ego_mm_dataset}. By organizing sequential video clips \{\text{Clip 1}, \text{Clip 2}, ..., \text{Clip N}\} and their corresponding narrations \{\text{Narration 1}, \text{Narration 2}, ..., \text{Narration N}\} in proper chronological order, we create comprehensive narration paragraphs that describe entire video sequences. 
We then employ a powerful \textbf{text-only language model}, i.e., GPT-4o, to generate diverse and confident QA pairs related to episodic memory based on these narration paragraphs. The language model is instructed to attach the index of the narration sentence upon which each QA pair is based. 
This indexing allows us to map each QA pair back to the corresponding time frames in the original videos, enabling the extraction of key frame information crucial for subsequent model training.

Applying this data engine to the extensive Ego4D dataset allows us to efficiently scale the creation of egocentric QA data. We partition the dataset into training and testing sets according to the official Ego4D episodic memory task.
The egocentric QA dataset provides more than 7 million QA samples in 938K multi-turn conversations.
The data encompasses videos of varying durations, ranging from 30 seconds to 1 hour, as illustrated in Figure~\ref{fig:mm_memory_stat}. 
To ensure comprehensive coverage and prevent bias towards shorter videos, we balance the number of conversations across different video lengths in training. 
This is one of the first large-scale egocentric QA datasets featuring videos of such extended ranges of duration.

Through these steps, our ``narration to egocentric QA" data engine addresses the scarcity of large-scale, high-quality egocentric QA data for egocentric scenes, and sets a solid foundation for building \modelname, a sophisticated egocentric MLLM, which we introduce in the following section.

\subsection{\modelname Model}

Our modeling goal is to develop an MLLM for handling egocentric videos, which are lengthy and rich in visual details. 
On the one hand, frame-level information is necessary to capture the full content of the video, as skipping frames during sampling could result in a significant loss of visual details. On the other hand, processing all visual tokens generated by the visual encoder is computationally challenging for the transformer model. 
For instance, if each image is encoded into 729 visual embeddings~(tokens), the total number of visual embeddings for a 300-frame video would be 218,700. However, most MLLMs are trained with a context length of less than 10,000 tokens~\citep{li2024llavaov}.
Taking these factors into account, we introduce the \modelname model, which is built for handling a large volume of egocentric video frames while maintaining manageable computational costs within the transformer backbone.
\modelname introduces an innovative Memory Pointer Prompting mechanism, which operates in two main steps: global glimpse and fallback. 
We will introduce the details of \modelname in the following sections.

\subsubsection{Visual and Textual Embedding}
Given an input video and the question, the first step is to embed them into visual and textual embeddings separately for later processing.
We begin by uniformly sampling the video into up to $N$ frames, where $N$ can be in the range of hundreds. Then, we extract per-frame visual feature maps from these frames using a robust vision encoder, SigLIP-so400m~\citep{zhai2023siglip}. Following the method outlined by \cite{li2024llavaov}, we apply a 2-layer MLP to project the visual feature maps to the LLM embedding space and use average pooling to reduce the height and width of the visual feature maps by a factor of two and flatten the height and width dimension, resulting in $N$ relatively high-resolution visual embeddings 
\{$\textbf{V}^i \in \mathbb{R}^{T\times C}$, $i \in [1, N]$\}
 where $T$ is the embedding length and $C$ is the embedding dimension. 
For the textual embedding, since we use Qwen2~\citep{yang2024qwen2} as the LLM, we use its tokenizer and embedding layer to transform the input text into textual embeddings. 
For question $q$, we denote the corresponding textual question embedding as \{$\textbf{E}_{\text{que}}^q \in \mathbb{R}^{T_q\times C}$, $q \in [1, Q]$\} where $Q$ is the total number of questions and $T_q$ is the embedding length of question $q$.

\begin{figure}[t]
    \centering
    \vspace{-37pt}
    \includegraphics[width=\textwidth,clip]
    {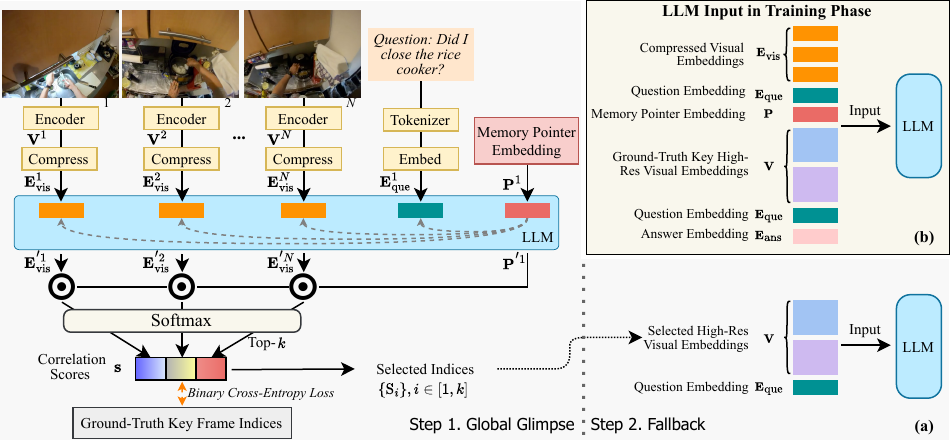}
    \vspace{-15pt}
    \caption{\textbf{(a)} Overview of the proposed Memory Pointer Prompting mechanism.
    Its inference consists of two steps: 
    (1) \emph{Global Glimpse}: We concatenate the compressed visual embeddings from all frames, denoted as $\mathbf{E}_{\text{vis}}^i$ for $i \in [1, N]$, with the question embeddings $\mathbf{E}_{\text{que}}^1$ and the memory pointer embedding $\mathbf{P}^1$. This combined embedding sequence is then input into the LLM. From the last layer, we extract embeddings and compute the dot product between the memory pointer embedding and all compressed visual embeddings to generate the correlation scores. The indices of the frames with the top $k$ scores are selected. During training, the correlation scores are supervised by ground-truth key frame indices via a binary cross-entropy loss.
    (2) \emph{Fallback}: The high-resolution visual embeddings corresponding to the selected indices are fed into the LLM along with the question embeddings for final processing and response generation.
    \textbf{(b)} Illustration of LLM input sequence during training.
    }
    \vspace{-10pt}
    \label{fig:pointer}
\end{figure}

\subsubsection{Memory Pointer Prompting}
\label{sec:pointer}
As processing all $N$ high-resolution visual embeddings with the LLM is computationally difficult, we propose to identify key visual embeddings in a question-aware manner and only send those selected embeddings to the subsequent LLM.
Inspired by previous works on Pointer Networks~\citep{vinyals2015pointer,merity2016pointer}, we propose a Memory Pointer Prompting mechanism, which is illustrated in Figure~\ref{fig:pointer}. 
Memory Pointer Prompting consists of two steps during inference: global glimpse and fallback.
In the global glimpse step, key visual embeddings are identified from all frame-level embeddings, guided by the context of the question. During the subsequent fallback step, the important visual embeddings are selected, and their higher-resolution versions are provided to the LLM transformer backbone for further processing and language response generation.

\paragraph{Global Glimpse Step.}
We begin by compressing the visual embeddings through average pooling along the embedding length dimension, resulting in a set of compressed visual embeddings \{$\textbf{E}^{i}_{\text{vis}} \in \mathbb{R}^{1\times C}, i \in [1, N]$\}. 
Next, we introduce a learnable memory pointer prompt embedding $\textbf{P} \in \mathbb{R}^{1\times C}$, 
duplicate it $Q$ times, yielding \{$\textbf{P}^{i} \in \mathbb{R}^{1\times C}, i \in [1, Q]$\}, 
and concatenate the embeddings as follows:
\begin{equation}
\small \nonumber
    [\textbf{E}^{1}_{\text{vis}}, \textbf{E}^{2}_{\text{vis}}, ..., \textbf{E}^{N}_{\text{vis}}, \textbf{E}_{\text{que}}^1, \textbf{P}^1].
\end{equation}
Here $Q=1$ as MLLMs generate answers for only one question at a time.
In this way, the question embedding is followed by a pointer embedding, which will be used to identify key visual embeddings with knowledge of the question embedding. The entire embedding sequence is then fed into the LLM, from which we obtain the output embedding sequence of the final layer:
\begin{equation}
\small \nonumber
    [\textbf{E}^{'1}_{\text{vis}}, \textbf{E}^{'2}_{\text{vis}}, ..., \textbf{E}^{'N}_{\text{vis}}, \textbf{E}_{\text{que}}^{'1}, \textbf{P}^{'1}].
\end{equation}
We extract and stack the processed visual embeddings \{$\textbf{E}^{'i}_{\text{vis}} \in \mathbb{R}^{1\times C}, i \in [1, N]$\} to obtain the matrix $\bm{E}_{\text{vis}} \in \mathbb{R}^{N\times C}$. 
We conduct a softmax dot product operation between  $\bm{E}_{\text{vis}}$ and $\textbf{P}^{'1}$: 
\begin{equation}
    \mathbf{s} = \text{Softmax} (\bm{E}_{\text{vis}} \cdot \textbf{P}^{'1\mathsf{T}}) \in \mathbb{R}^{N}.
\end{equation}
Here $\mathbf{s}$ is a correlation score vector indicating the correlation between the question and each frame.

\textbf{Balancing Exploration and Exploitation.}
Our approach to selecting key visual embeddings parallels the principles of Bayesian Optimization~\citep{frazier2018tutorial}, where the objective function is expensive to evaluate.
In such cases, it's important to balance exploration~(sampling in areas where the uncertainty is high) and exploitation~(sampling in areas where the surrogate model predicts high performance). However, relying solely on the aforementioned Memory Pointer Prompting may lead to overemphasizing certain areas of interest, potentially undermining the exploration process. To mitigate this issue, we introduce perturbations into the score distribution by incorporating a uniform sampling distribution.
The probability vector of uniform sampling can be written as:
\begin{equation}
\textbf{u}^i = 
\begin{cases} 
\alpha & \text{if } i \in \text{linspace}(0, N, k), \\
0 & \text{otherwise}.
\end{cases}
\end{equation}
Here $\alpha$ is an explore-exploit balancing parameter to adjust the probability distribution.
We overlap the probability vector of uniform sampling and score matrix $\textbf{s}$:
\begin{equation}
\textbf{s} \leftarrow \textbf{s} + \textbf{u}.
\end{equation}
We then identify the top-$k$ indices as the set $\{\ermS_i, i \in [1, k]\}$. In this way, we find the key visual embeddings in a question-aware manner. 

\paragraph{Fallback Step.}
During inference, as shown in Figure~\ref{fig:pointer},
with the set of indices $\{\ermS_i, i \in [1, k]\}$ for the selected visual embeddings, we now assemble the LLM input sequence as follows:
\begin{equation}
\small \nonumber
    [\underbrace{\textbf{V}^{\ermS_1}, \textbf{V}^{\ermS_2}, ..., \textbf{V}^{\ermS_k}}_{\text{Selected Top-}k\text{ Visual Embeedings}}, \textbf{E}_{\text{que}}^1].
\end{equation}
As previously introduced, $\textbf{V}^{\ermS_1}, \textbf{V}^{\ermS_2}, ..., \textbf{V}^{\ermS_k}$ denote the selected top-$k$ high-resolution visual embeddings, which provide more visual details than the compressed visual embeddings.
This new embedding sequence is fed into the LLM to generate the final language response.
In summary, the proposed Memory Pointer Prompting approach allows us to consider the full scope of video information while filtering out redundant data in the LLM transformer, ensuring computational efficiency. The new input serves as the final input of the LLM to generate the language response given the visual and textual information.

\paragraph{Training Procedure.}
Given the novel design of \modelname, its training procedure is different from popular MLLMs~\citep{liu2023llava}.
Specifically, let the answer embedding for question $q\in [1, Q]$ be denoted as $\mathbf{E}^q_{\text{ans}}$, 
then the input embedding sequence during the training process is represented as:
\begin{align}
\small \nonumber
    [\underbrace{\textbf{E}^{1}_{\text{\text{vis}}}, \textbf{E}^{2}_{\text{\text{vis}}}, ..., \textbf{E}^{N}_{\text{vis}}}_{\text{Compressed Visual Embeddings}}, \textbf{E}_{\text{que}}^1, \textbf{P}^1, ..., \textbf{E}_{\text{que}}^Q, \textbf{P}^Q, \underbrace{\textbf{V}^{\ermS_1}, \textbf{V}^{\ermS_2}, ..., \textbf{V}^{\ermS_k}}_{\text{Selected High-Res Visual Embeddings}}, \textbf{E}_{\text{que}}^1, \textbf{E}_{\text{ans}}^1, ..., \textbf{E}_{\text{que}}^Q, \textbf{E}_{\text{ans}}^Q].
\end{align}
We also provide a simplified illustration (where $Q=1$) of the input embedding sequence structure during training in Figure~\ref{fig:pointer}.
Here, we begin by inputting the compressed visual embeddings for all $N$ frames, followed by the question embedding and memory pointer embedding. Next, we integrate the $k$ selected high-resolution visual embeddings (based on the ground-truth key frame labels), and finally, incorporate both the question and answer embeddings.
Once the input sequence is prepared as outlined above, we can train \modelname similarly to traditional large language models.
The compressed visual embeddings, question embedding, and memory pointer embeddings used as prefixes do not contribute to the language cross-entropy loss.

When training on samples from our curated egocentric QA dataset where there are ground-truth key frame labels for each question, we compute the correlation score vector $\textbf{s}$ in the global glimpse step, and supervise it using a binary cross-entropy loss.
For training samples that lack ground-truth key frame labels, we omit the prefixes, which results in the traditional MLLM training process.

\section{Experiments}

In the experiment section, we will first present a new egocentric video understanding benchmark, specifically designed to assess episodic memory capabilities. Following this, we will perform comprehensive experiments to evaluate \modelname, utilizing both the newly introduced benchmark and existing public benchmarks.

\begin{table}[t]
  \centering
    \vspace{-20pt}
  \begin{minipage}{0.5\textwidth}
    \centering
    \caption{Distribution of videos and QA samples with different lengths.}
  \vspace{-10pt}
    \tablestyle{4pt}{1.2}\scriptsize
    \begin{tabular}{ c | cc | ccc | cc|c  c | c }  
    \toprule
    \multicolumn{1}{c|}{\multirow{1}{*}{\textbf{Class}}} & \multicolumn{2}{c|}{\textbf{Short}} & \multicolumn{3}{c|}{\textbf{Medium}} & \multicolumn{2}{c|}{\textbf{Long}} & \multicolumn{1}{c}{\textbf{Sum}}  \\ 
    \shline
    \textbf{Minutes} & 0.5-1 & 1-2 & 2-4 & 4-10 & 10-20 & 20-40 & 40-60 & -  \\
    \textbf{Videos} & 100 & 100 & 100 & 100 & 100 & 100 & 29 & 629 \\
    \textbf{QAs} & 500 & 498 & 987 & 997 & 1715 & 1792 & 537 & 7026 \\
      \bottomrule
    \end{tabular}
    \label{tab:ego_mem_stat}
  \end{minipage}
  \hspace{1cm} 
  \begin{minipage}{0.4\textwidth}
    \centering
    \caption{Distribution of correct options in MCQs.}
  \vspace{-10pt}
    \resizebox{.8\linewidth}{!}{
      \tablestyle{3pt}{1.2}\begin{tabular}{l| ccc ccc }
      \toprule
       \textbf{Option} & \multicolumn{1}{c}{\textbf{A}}  &  \textbf{B} & \textbf{C} & \textbf{D}    \\
      \shline
       \textbf{Counts} & 1776 & 1751 & 1770 & 1729 \\
      \bottomrule
      \end{tabular}
    }
    \label{tab:mcq_stat}
  \end{minipage}
  \label{tab:combined_subtables}
  \vspace{-10pt}
\end{table}

\subsection{\benchname Benchmark}

\begin{wrapfigure}{r}{0.5\textwidth}
\vspace{-45 pt}
  \begin{center}
     \includegraphics[width=0.5\textwidth]{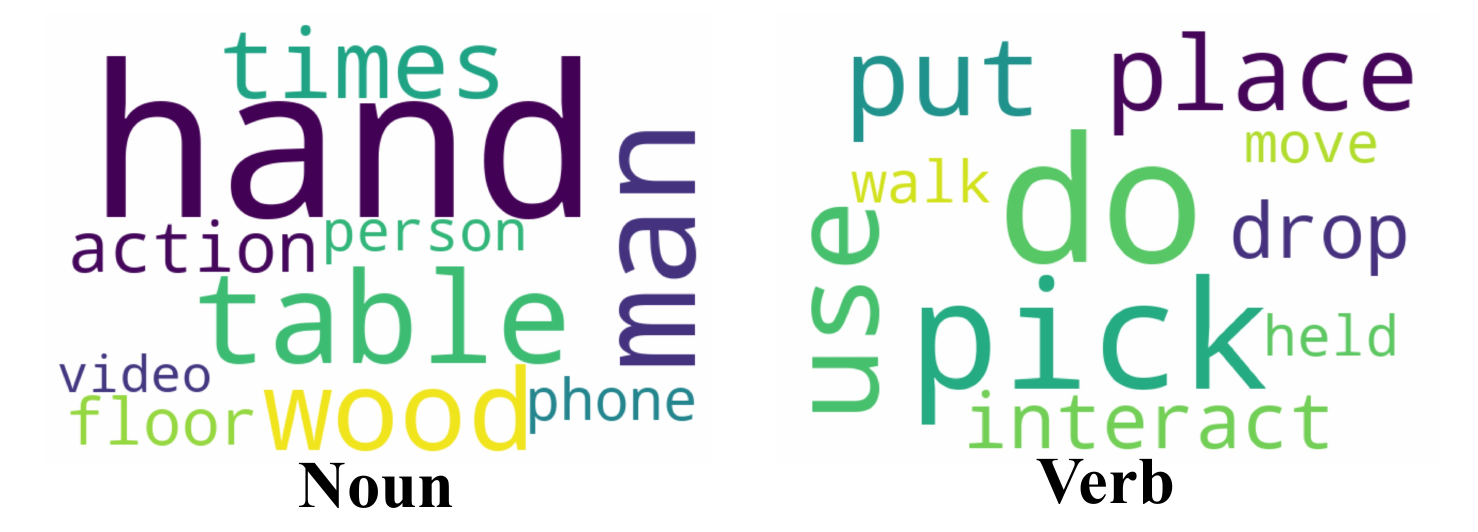} %
     \vspace{-18pt}
  \end{center}
  \vspace{-0.2cm}
   \caption{The most frequently occurring verbs and nouns in \benchname.}
  \label{fig:word_freq}
  \vspace{-0.3cm}
\end{wrapfigure}

To evaluate the performance of egocentric MLLMs, especially in terms of episodic memory ability, we propose a new benchmark called \benchname.
Specifically, we generate memory-related questions and answers from human-annotated narrations in the validation set of the Ego4D dataset.
To ensure diversity, for each video we only generate a limited number of questions.
We divide the videos into seven different length ranges: 0.5 to 1 min, 1 to 2 min, 2 to 4 min, 4 to 10 min, 10 to 20 min, 20 to 40 min, and 40 to 60 min. We aim to balance the number of samples in different video lengths. The distribution of videos and corresponding question-answer pairs (QAs) for each category is shown in Table~\ref{tab:ego_mem_stat}.  
Furthermore, we group these video lengths into three broader categories: short (0.5 to 2 min), medium (2 to 20 min), and long (20 to 60 min). In total, we collect 629 videos with 7,026 questions. 
The most frequently occurring verbs and nouns in the questions are visualized in Figure~\ref{fig:word_freq}.

\begin{figure}[t]
    \centering
    \vspace{-35pt}
    \includegraphics[width=\textwidth,clip]
    {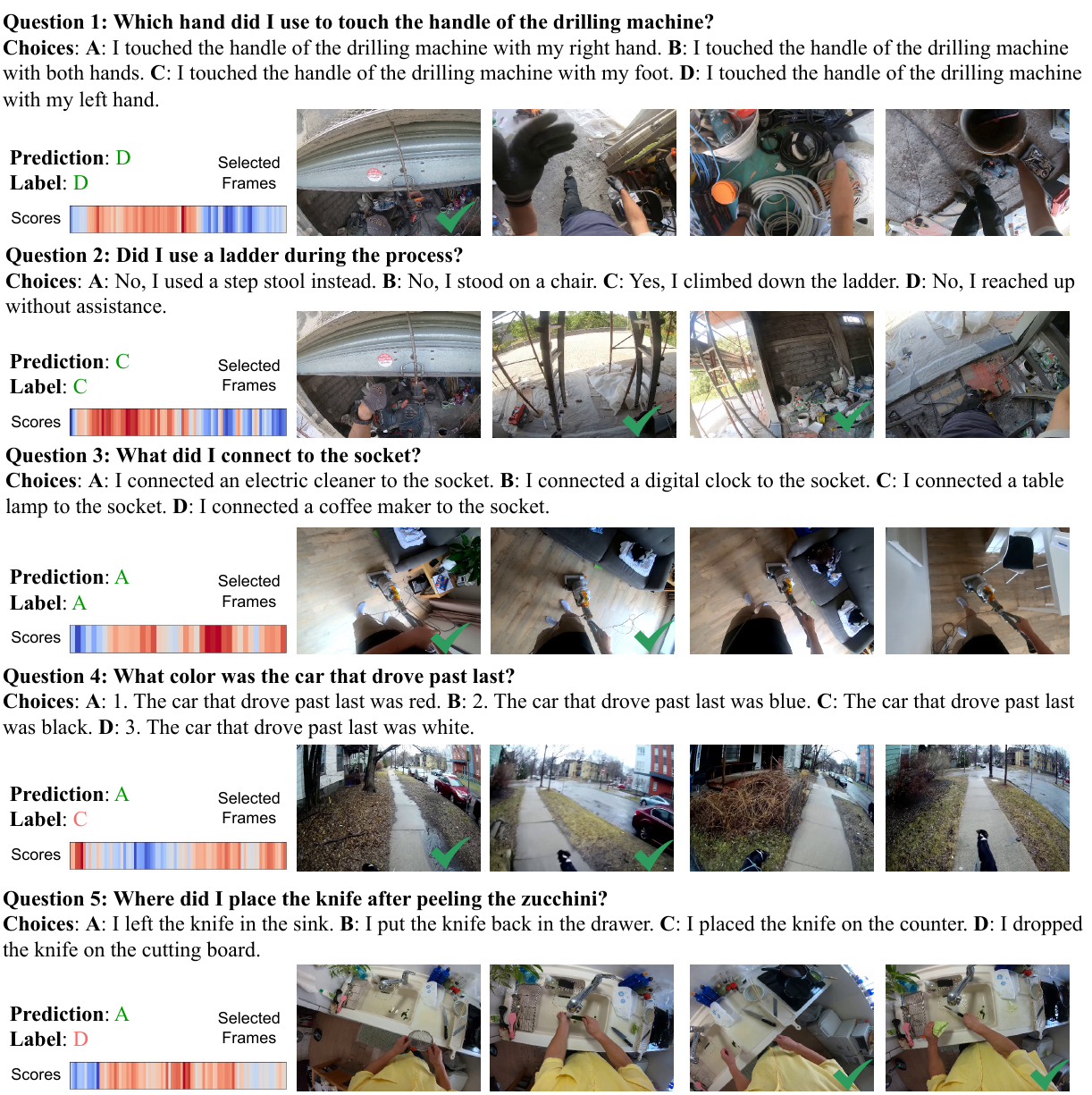}
    \vspace{-15pt}
    \caption{\benchname QAs visualization and prediction analysis of the global glimpse step. We find high consistency between the identified key frames and the questions, demonstrating the effectiveness of the proposed Memory Pointer Prompting method.
    The visualized correlation scores show distinct distributions for different questions given the same video, indicating its question-specific nature. The \textcolor{green}{\checkmark} indicates that the selected frames are relevant to the questions.
    }
    \vspace{-20pt}
    \label{fig:frame_selection}
\end{figure}

Since free-form answers are typically evaluated using a closed-source LLM as a judge, the evaluation can be inconsistent and subject to significant variance, especially due to model version updates. To ensure more reliable, standardized, and consistent performance evaluation, we convert the free-form answers into multiple-choice questions~(MCQs), which helps reduce score instability. In practice, based on the free-form answer, we instruct ChatGPT to generate three additional choices that are plausible but incorrect, considering the original question and answer. We then randomize the order of these choices to achieve a uniform distribution of correct options, as shown in Table~\ref{tab:mcq_stat}, to minimize bias in option placement.
We visualize some randomly sampled examples in Figure~\ref{fig:frame_selection}.

\subsection{Experimental Setup}
\label{sec:exp_setup}

\textbf{Training Data.}
We employ a joint image-video supervised fine-tuning (SFT) strategy. To enhance the model's capability in understanding a broader range of visual data, we combine our egocentric QA dataset with a variety of multimodal datasets.
We curate an SFT dataset mixture consisting of
our egocentric QA dataset, 
Ego4D narration dataset~\citep{grauman2022ego4d},
LLaVA-NeXT SFT collection~(including ChartQA~\citep{masry2022chartqa}, AI2D~\citep{hiippala2021ai2d}, DocVQA~\citep{mathew2021docvqa}, DVQA~\citep{kafle2018dvqa}, COCO~\citep{lin2014COCO}), ShareGPT4V~\citep{chen2023sharegpt4v}, synthdog-en~\citep{kim2021donut}), ShareGPT-4o~\citep{chen2023internvl}, ALLaVA instruct~\citep{chen2024allava}, ShareGPT4Video~\citep{chen2024sharegpt4video}, sherlock~\citep{hesselhwang2022abduction}, ScienceQA~\citep{scienceqa}, NExT-QA~\citep{xiao2021next}, and ActivityNet-QA~\citep{yu2019activitynet}.

\textbf{Implementation Details.}
The model is trained for 1 epoch with a base learning rate of $1\times 10^{-5}$, using a cosine scheduler. The batch size is set to 128. We sample a maximum of 300 frames~($N=300$) and select 32 visual embeddings in the proposed memory pointer prompting mechanism. By default, we set the explore-exploit balancing parameter $\alpha$ to $0.1$. Greedy decoding is used in generation.

\textbf{Pretrained Models.}
Our \modelname model is initialized from LLaVA-OV 7B~\citep{li2024llavaov}, a state-of-the-art MLLM known for its good performance on general multimodal understanding tasks.
Following the same architecture, we use the SigLip-so400M ViT~\citep{zhai2023siglip} as the visual encoder for embedding video frames and Qwen2-7B~\citep{yang2024qwen2} as the LLM architecture.

\subsection{Main Results}

We first conduct experiments on our \benchname benchmark, primarily comparing three models: LLaVA-OV~\citep{li2024llavaov}, its fine-tuned version using our \modelname SFT data mixture (referred to as ``Ego SFT"), and our \modelname model, which incorporates the proposed Memory Pointer Prompting mentioned in Section~\ref{sec:pointer}. We show the \benchname accuracy in the first row of Table~\ref{tab:ego_mem_main_results}.
We observe a significant improvement in the model's performance on egocentric QAs after training on our MM-Ego data mixture, attributed to the rich egocentric knowledge provided by our curated egocentric QA training data. Moreover, leveraging the \modelname model architecture further enhances performance, thanks to the effective Memory Pointer Prompting mechanism.

However, we notice that the original overall performance metrics are higher than anticipated, raising curiosity about the extent to which language bias contributes to the models' accuracy. To answer this question, we conduct additional experiments aimed at eliminating these language biases.
Specifically, we test the three model variants on the \benchname benchmark without any visual inputs, identifying questions that could be correctly answered without videos as ``language-biased questions". 
Then, we evaluate the models' performance on the subset of the benchmark without language-biased questions.
For fairness, we apply this debiasing process across all three models so that they are evaluated on the same sets of data. We calculate the mean accuracy of the debiased variants, referred to as the ``Mean Debiased Accuracy (MDA)". The results are presented in Table~\ref{tab:ego_mem_main_results}.

As expected, after removing the language-biased questions, the accuracy of all three models drops significantly to a more reasonable level. 
The performance decline is notably more pronounced in the ``Medium" and ``Long" classes compared to the ``Short" class.
For example, the average accuracy of LLaVA-OV across the three classes (short, medium, and long) drops from 65.45 to 47.32. The decrease in the ``Short" class is 17.04, in the ``Medium" class is 17.93, and in the ``Long" class is 19.43. 
Despite this, we still observe improvements in MDA after training with SFT data generated by our \modelname data engine~(\textbf{+8.65}) and applying our Memory Pointer Prompting method~(\textbf{+13.95}).
These results demonstrate the effectiveness of our approach even after considering language bias.

\begin{table}[t]
\vspace{-20pt}
  \centering
   \caption{{Performance comparison and language bias analysis of different models on the \benchname benchmark.} Our \modelname model demonstrates the best performance both before and after excluding the language bias of different models.}
  \vspace{-10pt}
  \tablestyle{2.5pt}{1.2}\scriptsize\begin{tabular}{l| cccc | cccc | cccc }
  \toprule
 \multicolumn{1}{c|}{\multirow{2}{*}{\textbf{Method}} }  & \multicolumn{4}{c|}{\textbf{LLaVA-OV~\citep{li2024llavaov}}} &  \multicolumn{4}{c|}{\textbf{Ego SFT}} & \multicolumn{4}{c}{\textbf{\modelname}} \\
   & Short & Medium & Long & Avg & Short & Medium & Long & Avg & Short & Medium & Long & Avg  \\
  \shline
  Original & 70.24 & 64.94 & 61.19 & 65.45                              & 79.06 & 76.34 & 73.51 & 76.30 & {79.96} & 79.64 & 79.09 & \textbf{79.56}  \\
  Exclude \textit{LLaVA-OV} Bias & 56.44 & 49.64 & 44.83 & 50.30 & 66.37 & 64.15 & 60.03 & 63.52 & 71.97 & 70.68 & 68.15 & \textbf{70.26} \\
  Exclude \textit{Ego SFT} Bias & 55.75 & 49.27 & 45.21 & 50.08 & 61.73 & 59.59 & 54.50 & 58.61 & 67.70 & 66.33 & 63.89 & \textbf{65.97} \\
  Exclude \textit{\modelname} Bias & 47.41 & 42.11 & 35.22 & 41.58 & 50.60  & 46.39 & 40.38 & 45.79 & 49.80 & 49.11 & 43.81 & \textbf{47.58}\\
   \rowcolor{gray!15}
  Mean Debiased Accuracy (MDA) & 53.20&	47.01&	41.76 & {47.32} & 59.56 & 56.71 & 51.64 & {55.97} & 63.16 & 62.04 & 58.62 & \textbf{61.27}  \\
  \bottomrule
  \end{tabular}
  \vspace{-10pt}
\label{tab:ego_mem_main_results}
\end{table}

\begin{table}[t]
\vspace{-10pt}
  \centering
   \caption{Comparison with state-of-the-art video MLLMs. \modelname shows strong performance on egocentric understanding and competitive performance on Internet video understanding.}
  \vspace{-10pt}
  \tablestyle{3pt}{1.2}\scriptsize\begin{tabular}{l | cccc |c|cc  c |c  cccc }
  \toprule
  \multicolumn{1}{c|}{\multirow{2}{*}{\textbf{Method}} }  & \multicolumn{4}{c|}{\textbf{\benchname (MDA)}} & \textbf{EgoSchema} &  \multicolumn{4}{c}{\textbf{Video-MME (w/o subs)}} \\
  & Short & Medium & Long & Avg & Full & Short & Medium & Long & Entire \\
  \shline
  GPT-4o & \textbf{64.31} & 59.47 & 57.65 & 60.48  & \textbf{72.2} & \textbf{80.00} & \textbf{70.30} & \textbf{65.30} & \textbf{71.90}\\
  LLaVA-NeXT-Video-7B-DPO~\citep{zhang2024llavanextvideo} & 30.38 & 25.95 & 21.49 & 25.94 & - & - & - & - & - \\
  LLaVA-NeXT-Video-32B-Qwen~\citep{zhang2024llavanextvideo} & 43.78 & 33.76 & 31.04 & 36.19 & 60.85 & - & - & - & 60.20  \\
  LLaVA-OV 7B~\citep{li2024llavaov} & 53.20 & 47.01& 41.76 & 47.32 & 60.10 & 69.30 	 & 56.00 & 	49.40 &	58.30 \\
  \rowcolor{gray!15}
  \modelname (ours) & 63.16 & \textbf{62.04} & \textbf{58.62} & \textbf{61.27} & {69.03} & 67.60 & 55.70 & 47.80 & 57.00 \\
  \bottomrule
  \end{tabular}
  \vspace{-10pt}
\label{tab:public_bench}
\end{table}

\begin{table}[h]
  \centering
  \caption{MDA on \benchname when inferring with different numbers of frames. Our \modelname model shows a smaller relative drop on average when decreasing the number of sampled frames.}
  \vspace{-10pt}
    \resizebox{1.\linewidth}{!}{
  \tablestyle{3pt}{1.2}
  \scriptsize
  \begin{tabular}{c|ccc|ccc|ccc|cccc}
  \toprule
  \multirow{2}{*}{\textbf{Frames}} & \multicolumn{3}{c|}{\textbf{Short}} & \multicolumn{3}{c|}{\textbf{Medium}} & \multicolumn{3}{c|}{\textbf{Long}} & \multicolumn{3}{c}{\textbf{Avg}} \\
  &  LLaVA-OV & Ego SFT & \modelname & LLaVA-OV & Ego SFT  & \modelname & LLaVA-OV & Ego SFT  & \modelname & LLaVA-OV & Ego SFT  & \modelname  \\
  \shline
  \textbf{32} & 53.20 & 59.56 & 63.16 & 47.01 & 56.71 & 62.04 & 41.76 & 51.64 & 58.62 & 47.32 & 55.97 & 61.27 \\
  \textbf{16} & 52.68 & 60.45 & 63.82 & 46.37 & 55.99 & 60.81 & 40.12 & 51.15 & 58.16 & 46.39 & 55.86 & 60.93 \\
  \textbf{8}  & 50.76 & 59.59 & 62.22 & 44.82 & 54.55 & 58.23 & 39.41 & 49.11 & 55.19 & 44.99 & 54.42 & 58.55 \\
  \textbf{4}  & 50.43 & 55.36 & 62.30 & 42.54 & 52.08 & 58.44 & 38.88 & 48.40 & 54.65 & 43.95 & 51.95 & 58.46 \\
  \rowcolor{gray!15}
  \textbf{Rel. Diff} & 5.20\% & 7.07\% & \textbf{1.36\%} & 9.49\% & 8.16\% & \textbf{5.81\%} & 6.89\% & \textbf{6.26\%} &{ 6.77\%} & 7.12\% & 7.19\% & \textbf{4.59\%} \\
  \bottomrule
  \end{tabular}}
  \vspace{-15pt}
  \label{tab:infer_frames}
\end{table}

To better understand the capability of \modelname, 
we compare its performance with state-of-the-art video MLLMs on \benchname and prevalent large-scale video QA benchmarks, including the long egocentric video understanding benchmark EgoSchema~\citep{mangalam2023egoschema} and the Internet-video-based long-video understanding benchmark Video-MME~\citep{fu2024videomme}.
The results are shown in Table~\ref{tab:public_bench}.
On \benchname, GPT-4o is evaluated using 32 uniformly sampled frames from the videos, while other models follow their respective official inference settings. The MDA on \benchname is computed using the debiased subsets used in Table~\ref{tab:ego_mem_main_results}.
Notably, \modelname exhibits the highest performance on \benchname, particularly in the `Medium' and `Long' classes. On the EgoSchema benchmark, our model achieves a substantial performance gain of \textbf{+8.18} over the previous state-of-the-art open-source model (``LLaVA-NeXT-Video-32B-Qwen''), underscoring the effectiveness of both our data and model design for egocentric understanding.
Additionally, on the challenging Internet video understanding Video-MME benchmark, our model is on par with the leading model of similar parameter size although our data mixture is less diverse compared with~\citep{li2024llavaov}.
These results showcase \modelname's capability in egocentric video understanding while preserving its general video comprehension abilities.

\begin{wrapfigure}{r}{0.32\textwidth}
\vspace{-2cm}
  \begin{center}
     \includegraphics[width=0.32\textwidth]{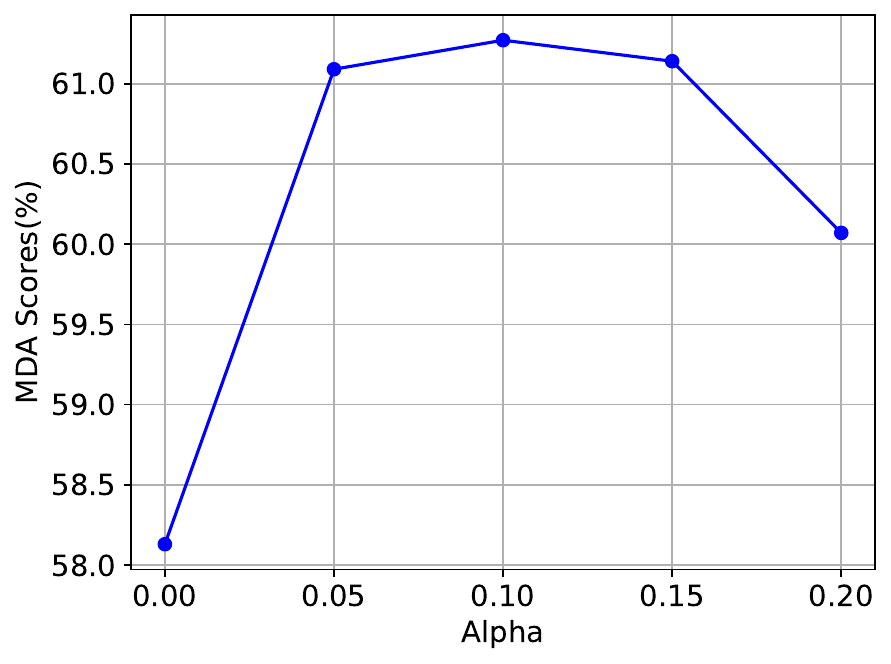}
     \vspace{-18pt}
  \end{center}
   \caption{MDA scores with different $\alpha$ values for explore-and-exploit balancing.}
  \label{fig:alpha}
  \vspace{-4pt}
\end{wrapfigure}

\subsection{Model Analysis}
\paragraph{Quantitative Analysis of Different Numbers of Frames.}
To evaluate the influence of sampling different numbers of frames for different models, we calculate the mean debiased accuracy~(MDA) in Table~\ref{tab:infer_frames}. The relative performance drop from sampling 32 frames to sampling 4 frames is also calculated. 
As expected, all models exhibit a decrease in performance with fewer sampled frames. 
Notably, \modelname exhibits a smaller average performance drop when the number of frames is reduced due to its ability to identify key frames given lower computational budget.
The relative performance drop in the short category is considerably smaller compared to the medium and long categories, likely because shorter videos require fewer frames to comprehend.

\paragraph{Qualitative Analysis of Memory Pointer Prompting.}
In Figure~\ref{fig:frame_selection}, we present a qualitative analysis of the accuracy of Memory Pointer Prompting on \benchname. We randomly select samples and visualize the key frames identified by the global glimpse step in Memory Pointer Prompting. The results show a strong alignment between the questions and the selected frames. In failure cases, we observe that the issues are often due to the ambiguity of the questions, causing the model to struggle with accurately localizing the key visual embeddings.
Furthermore, the visualized correlation scores during the global glimpse step show distinct patterns across various videos and questions, confirming its effectiveness in selecting key visual embeddings tailored to the specific questions.

\paragraph{Quantitative Analysis of Explore-Exploit Balancing Parameter $\alpha$.}
As discussed in Section~\ref{sec:pointer}, we design an explore-exploit balancing parameter $\alpha$ to fuse the uniform distribution and the sampling probability computed by Memory Pointer Prompting.
We illustrate \modelname's performance with varying values of $\alpha$ in Figure~\ref{fig:alpha}. The results show that $\alpha=0.1$ achieves the best performance, while larger or smaller values of $\alpha$ tend to either over-explore or over-exploit.

\begin{figure}[t]
    \centering
    \vspace{-30pt}
    \includegraphics[width=\textwidth,clip]
    {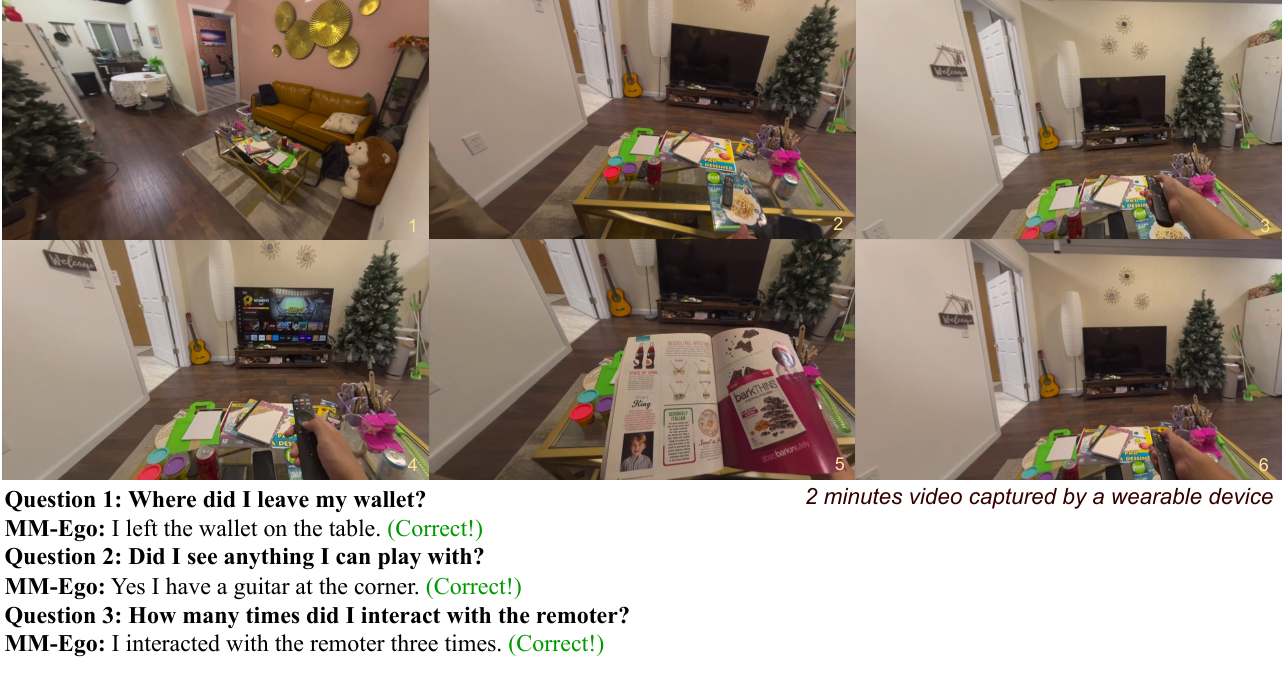}
    \vspace{-25pt}
    \caption{Real-world conversation examples generated by \modelname. The input is a 2-minute long egocentric video recorded using a camera on an off-the-shelf wearable device. \modelname can accurately identify key visual details and provide correct answers to the user's memory-related questions.
    }
    \vspace{-10pt}
    \label{fig:conversation2}
\end{figure}

\paragraph{Conversation Examples by \modelname.}
In Figure~\ref{fig:conversation2}, we show a real-world demo of \modelname, where the input video is a 2-minute long egocentric video captured by a camera on an off-the-shelf wearable device (this video is not used in our dataset). \modelname is able to correctly answer the episodic memory-related questions given the egocentric video, despite the difference in data domain.

\section{Related Work}
\label{sec:related_work}

\paragraph{Multimodal Large Language Models.} Recent advancements in Large Language Models~\citep{OpenAI_ChatGPT,touvron2023llama} have sparked significant interest in developing Multimodal Large Language Models (MLLMs) that combine the language understanding capabilities of LLMs with multi-modal perception abilities~\citep{alayrac2022flamingo,instructblip,zhu2023minigpt,mckinzie2024mm1}. 
For video-based MLLMs, most works follow a structure akin to image-based MLLMs. 
To handle the large volume of video frames, some methods reduce the number of frames \citep{damonlpsg2023videollama,wang2024internvideo2,Maaz2023VideoChatGPT,xu2024slowfast}, which results in the loss of many visual details. Others extend the LLMs' context length by employing parallel techniques \citep{xue2024longvila}, but this often leads to low training efficiency.
Unlike these approaches, our method preserves global awareness of the entire video, allows for attention to visual details, and is efficiently trainable.

\paragraph{Egocentric Video Understanding.}
While the growing field of egocentric video understanding is still in its infancy, there have been many influential works. For a comprehensive overview of egocentric vision please refer to \cite{plizzari2024outlook}.
On the data/benchmark side, 
representative works include Ego4D \citep{grauman2022ego4d}, Ego-Exo4D \citep{grauman2024ego}, and EPIC-KITCHENS-100 \citep{Damen2018EPICKITCHENS}.
When also considering language, prior work on egocentric video-language benchmarks include  QaEgo4D \citep{barmann2022qaego4d} and EgoSchema \citep{mangalam2023egoschema}.
For understanding long egocentric videos, prior modeling efforts include GroundVQA \citep{di2024groundvqa}, Encode-Store-Retrieve \citep{shen2023encode}, and R-VLM \citep{xu2023retrieval}.
However, most previous works focus on classic video understanding tasks such as activity recognition and temporal grounding, and hence they do not involve a large language model for complex understanding.
In contrast, we propose to develop an MLLM to tackle comprehensive egocentric video understanding.

\section{Conclusion}

In this paper, we make three key contributions towards the development of egocentric foundation models: the creation of a large-scale egocentric QA training dataset, the introduction of a novel model designed for effective long egocentric video comprehension, and the establishment of the \benchname benchmark for assessing models' ability to capture visual details from egocentric videos. We hope that these efforts will benefit further research on egocentric MLLMs.

\section*{Ethics Statement}
Our proposed method does not involve the creation or introduction of any new video content. All generated data is derived from publicly available, privacy-protected datasets~\citep{grauman2022ego4d}. 
The data is intended exclusively for academic research purposes and will not be used for any commercial applications. 
We have adhered to ethical standards by ensuring that no private or sensitive data has been used or compromised.

\section*{Reproducibility Statement}
We provide a detailed explanation of the data synthesis process in our data engine in Section~\ref{sec:egomm_dataset}.
We also elaborate on our model design in Section~\ref{sec:pointer}.
Additionally, we outline the implementation details, including the training hyperparameters in Section~\ref{sec:exp_setup}.

\section*{Acknowledgment}
This research is supported in part by the Early Career Scheme of the Research Grants Council (RGC) of the Hong Kong SAR under grant No. 26202321 and SAIL Research Project. 

\bibliography{main}

\begin{thebibliography}{61}
\providecommand{\natexlab}[1]{#1}
\providecommand{\url}[1]{\texttt{#1}}
\expandafter\ifx\csname urlstyle\endcsname\relax
  \providecommand{\doi}[1]{doi: #1}\else
  \providecommand{\doi}{doi: \begingroup \urlstyle{rm}\Url}\fi

\bibitem[Alayrac et~al.(2022)Alayrac, Donahue, Luc, Miech, Barr, Hasson, Lenc, Mensch, Millican, Reynolds, et~al.]{alayrac2022flamingo}
Jean-Baptiste Alayrac, Jeff Donahue, Pauline Luc, Antoine Miech, Iain Barr, Yana Hasson, Karel Lenc, Arthur Mensch, Katherine Millican, Malcolm Reynolds, et~al.
\newblock Flamingo: a visual language model for few-shot learning.
\newblock In \emph{NeurIPS}, 2022.

\bibitem[B{\"a}rmann \& Waibel(2022)B{\"a}rmann and Waibel]{barmann2022qaego4d}
Leonard B{\"a}rmann and Alex Waibel.
\newblock Where did i leave my keys?-episodic-memory-based question answering on egocentric videos.
\newblock In \emph{CVPR-W}, 2022.

\bibitem[Cai et~al.(2016)Cai, Kitani, and Sato]{cai2016understanding}
Minjie Cai, Kris~M Kitani, and Yoichi Sato.
\newblock Understanding hand-object manipulation with grasp types and object attributes.
\newblock In \emph{Robotics: Science and Systems}, volume~3, 2016.

\bibitem[Cartas et~al.(2017)Cartas, Mar{\'\i}n, Radeva, and Dimiccoli]{cartas2017recognizing}
Alejandro Cartas, Juan Mar{\'\i}n, Petia Radeva, and Mariella Dimiccoli.
\newblock Recognizing activities of daily living from egocentric images.
\newblock In \emph{Pattern Recognition and Image Analysis}, 2017.

\bibitem[Chandrasegaran et~al.(2024)Chandrasegaran, Gupta, Hadzic, Kota, He, Eyzaguirre, Durante, Li, Wu, and Fei-Fei]{chandrasegaran2024hourvideo}
Keshigeyan Chandrasegaran, Agrim Gupta, Lea~M Hadzic, Taran Kota, Jimming He, Crist{\'o}bal Eyzaguirre, Zane Durante, Manling Li, Jiajun Wu, and Li~Fei-Fei.
\newblock Hourvideo: 1-hour video-language understanding.
\newblock \emph{arXiv}, 2024.

\bibitem[Chen et~al.(2024{\natexlab{a}})Chen, Chen, Zhang, Chen, Wu, Zhang, Chen, Li, Wan, and Wang]{chen2024allava}
Guiming~Hardy Chen, Shunian Chen, Ruifei Zhang, Junying Chen, Xiangbo Wu, Zhiyi Zhang, Zhihong Chen, Jianquan Li, Xiang Wan, and Benyou Wang.
\newblock Allava: Harnessing gpt4v-synthesized data for a lite vision-language model.
\newblock \emph{arXiv}, 2024{\natexlab{a}}.

\bibitem[Chen et~al.(2023{\natexlab{a}})Chen, Li, Dong, Zhang, He, Wang, Zhao, and Lin]{chen2023sharegpt4v}
Lin Chen, Jisong Li, Xiaoyi Dong, Pan Zhang, Conghui He, Jiaqi Wang, Feng Zhao, and Dahua Lin.
\newblock Sharegpt4v: Improving large multi-modal models with better captions.
\newblock \emph{arXiv}, 2023{\natexlab{a}}.

\bibitem[Chen et~al.(2024{\natexlab{b}})Chen, Wei, Li, Dong, Zhang, Zang, Chen, Duan, Lin, Tang, Yuan, Qiao, Lin, Zhao, and Wang]{chen2024sharegpt4video}
Lin Chen, Xilin Wei, Jinsong Li, Xiaoyi Dong, Pan Zhang, Yuhang Zang, Zehui Chen, Haodong Duan, Bin Lin, Zhenyu Tang, Li~Yuan, Yu~Qiao, Dahua Lin, Feng Zhao, and Jiaqi Wang.
\newblock Sharegpt4video: Improving video understanding and generation with better captions.
\newblock \emph{arXiv}, 2024{\natexlab{b}}.

\bibitem[Chen et~al.(2023{\natexlab{b}})Chen, Wu, Wang, Su, Chen, Xing, Zhong, Zhang, Zhu, Lu, Li, Luo, Lu, Qiao, and Dai]{chen2023internvl}
Zhe Chen, Jiannan Wu, Wenhai Wang, Weijie Su, Guo Chen, Sen Xing, Muyan Zhong, Qinglong Zhang, Xizhou Zhu, Lewei Lu, Bin Li, Ping Luo, Tong Lu, Yu~Qiao, and Jifeng Dai.
\newblock Internvl: Scaling up vision foundation models and aligning for generic visual-linguistic tasks.
\newblock \emph{arXiv}, 2023{\natexlab{b}}.

\bibitem[Cheng et~al.(2024)Cheng, Guo, Wu, Fang, Li, Liu, and Liu]{cheng2024egothink}
Sijie Cheng, Zhicheng Guo, Jingwen Wu, Kechen Fang, Peng Li, Huaping Liu, and Yang Liu.
\newblock Egothink: Evaluating first-person perspective thinking capability of vision-language models.
\newblock In \emph{CVPR}, 2024.

\bibitem[Dai et~al.(2023)Dai, Li, Li, Tiong, Zhao, Wang, Li, Fung, and Hoi]{instructblip}
Wenliang Dai, Junnan Li, Dongxu Li, Anthony Tiong, Junqi Zhao, Weisheng Wang, Boyang Li, Pascale Fung, and Steven Hoi.
\newblock Instructblip: Towards general-purpose vision-language models with instruction tuning.
\newblock \emph{arXiv}, 2023.

\bibitem[Damen et~al.(2018)Damen, Doughty, Farinella, Fidler, Furnari, Kazakos, Moltisanti, Munro, Perrett, Price, and Wray]{Damen2018EPICKITCHENS}
Dima Damen, Hazel Doughty, Giovanni~Maria Farinella, Sanja Fidler, Antonino Furnari, Evangelos Kazakos, Davide Moltisanti, Jonathan Munro, Toby Perrett, Will Price, and Michael Wray.
\newblock Scaling egocentric vision: The epic-kitchens dataset.
\newblock In \emph{ECCV}, 2018.

\bibitem[Di \& Xie(2024)Di and Xie]{di2024groundvqa}
Shangzhe Di and Weidi Xie.
\newblock Grounded question-answering in long egocentric videos.
\newblock In \emph{CVPR}, 2024.

\bibitem[Frazier(2018)]{frazier2018tutorial}
Peter~I Frazier.
\newblock A tutorial on bayesian optimization.
\newblock \emph{arXiv}, 2018.

\bibitem[Fu et~al.(2024)Fu, Dai, Luo, Li, Ren, Zhang, Wang, Zhou, Shen, Zhang, et~al.]{fu2024videomme}
Chaoyou Fu, Yuhan Dai, Yondong Luo, Lei Li, Shuhuai Ren, Renrui Zhang, Zihan Wang, Chenyu Zhou, Yunhang Shen, Mengdan Zhang, et~al.
\newblock Video-mme: The first-ever comprehensive evaluation benchmark of multi-modal llms in video analysis.
\newblock \emph{arXiv}, 2024.

\bibitem[Grauman et~al.(2022)Grauman, Westbury, Byrne, Chavis, Furnari, Girdhar, Hamburger, Jiang, Liu, Liu, Martin, Nagarajan, Radosavovic, Ramakrishnan, Ryan, Sharma, Wray, Xu, Xu, Zhao, Bansal, Batra, Cartillier, Crane, Do, Doulaty, Erapalli, Feichtenhofer, Fragomeni, Fu, Gebreselasie, Gonzalez, Hillis, Huang, Huang, Jia, Khoo, Kolar, Kottur, Kumar, Landini, Li, Li, Li, Mangalam, Modhugu, Munro, Murrell, Nishiyasu, Price, Puentes, Ramazanova, Sari, Somasundaram, Southerland, Sugano, Tao, Vo, Wang, Wu, Yagi, Zhao, Zhu, Arbelaez, Crandall, Damen, Farinella, Fuegen, Ghanem, Ithapu, Jawahar, Joo, Kitani, Li, Newcombe, Oliva, Park, Rehg, Sato, Shi, Shou, Torralba, Torresani, Yan, and Malik]{grauman2022ego4d}
Kristen Grauman, Andrew Westbury, Eugene Byrne, Zachary Chavis, Antonino Furnari, Rohit Girdhar, Jackson Hamburger, Hao Jiang, Miao Liu, Xingyu Liu, Miguel Martin, Tushar Nagarajan, Ilija Radosavovic, Santhosh~Kumar Ramakrishnan, Fiona Ryan, Jayant Sharma, Michael Wray, Mengmeng Xu, Eric~Zhongcong Xu, Chen Zhao, Siddhant Bansal, Dhruv Batra, Vincent Cartillier, Sean Crane, Tien Do, Morrie Doulaty, Akshay Erapalli, Christoph Feichtenhofer, Adriano Fragomeni, Qichen Fu, Abrham Gebreselasie, Cristina Gonzalez, James Hillis, Xuhua Huang, Yifei Huang, Wenqi Jia, Weslie Khoo, Jachym Kolar, Satwik Kottur, Anurag Kumar, Federico Landini, Chao Li, Yanghao Li, Zhenqiang Li, Karttikeya Mangalam, Raghava Modhugu, Jonathan Munro, Tullie Murrell, Takumi Nishiyasu, Will Price, Paola~Ruiz Puentes, Merey Ramazanova, Leda Sari, Kiran Somasundaram, Audrey Southerland, Yusuke Sugano, Ruijie Tao, Minh Vo, Yuchen Wang, Xindi Wu, Takuma Yagi, Ziwei Zhao, Yunyi Zhu, Pablo Arbelaez, David Crandall, Dima Damen, Giovanni~Maria
  Farinella, Christian Fuegen, Bernard Ghanem, Vamsi~Krishna Ithapu, C.~V. Jawahar, Hanbyul Joo, Kris Kitani, Haizhou Li, Richard Newcombe, Aude Oliva, Hyun~Soo Park, James~M. Rehg, Yoichi Sato, Jianbo Shi, Mike~Zheng Shou, Antonio Torralba, Lorenzo Torresani, Mingfei Yan, and Jitendra Malik.
\newblock Ego4d: Around the world in 3,000 hours of egocentric video.
\newblock In \emph{CVPR}, 2022.

\bibitem[Grauman et~al.(2024)Grauman, Westbury, Torresani, Kitani, Malik, Afouras, Ashutosh, Baiyya, Bansal, Boote, et~al.]{grauman2024ego}
Kristen Grauman, Andrew Westbury, Lorenzo Torresani, Kris Kitani, Jitendra Malik, Triantafyllos Afouras, Kumar Ashutosh, Vijay Baiyya, Siddhant Bansal, Bikram Boote, et~al.
\newblock Ego-exo4d: Understanding skilled human activity from first-and third-person perspectives.
\newblock In \emph{Proceedings of the IEEE/CVF Conference on Computer Vision and Pattern Recognition}, pp.\  19383--19400, 2024.

\bibitem[Hessel et~al.(2022)Hessel, Hwang, Park, Zellers, Bhagavatula, Rohrbach, Saenko, and Choi]{hesselhwang2022abduction}
Jack Hessel, Jena~D Hwang, Jae~Sung Park, Rowan Zellers, Chandra Bhagavatula, Anna Rohrbach, Kate Saenko, and Yejin Choi.
\newblock {The Abduction of Sherlock Holmes: A Dataset for Visual Abductive Reasoning}.
\newblock In \emph{ECCV}, 2022.

\bibitem[Hiippala et~al.(2021)Hiippala, Alikhani, Haverinen, Kalliokoski, Logacheva, Orekhova, Tuomainen, Stone, and Bateman]{hiippala2021ai2d}
Tuomo Hiippala, Malihe Alikhani, Jonas Haverinen, Timo Kalliokoski, Evanfiya Logacheva, Serafina Orekhova, Aino Tuomainen, Matthew Stone, and John~A Bateman.
\newblock Ai2d-rst: A multimodal corpus of 1000 primary school science diagrams.
\newblock \emph{Language Resources and Evaluation}, 55:\penalty0 661--688, 2021.

\bibitem[Hochstein \& Ahissar(2002)Hochstein and Ahissar]{hochstein2002view}
Shaul Hochstein and Merav Ahissar.
\newblock View from the top: Hierarchies and reverse hierarchies in the visual system.
\newblock \emph{Neuron}, 36\penalty0 (5):\penalty0 791--804, 2002.

\bibitem[Huang et~al.(2024)Huang, Chen, Xu, Zhang, Yang, Pei, Zhang, Dong, Wang, Wang, et~al.]{huang2024egoexolearn}
Yifei Huang, Guo Chen, Jilan Xu, Mingfang Zhang, Lijin Yang, Baoqi Pei, Hongjie Zhang, Lu~Dong, Yali Wang, Limin Wang, et~al.
\newblock Egoexolearn: A dataset for bridging asynchronous ego-and exo-centric view of procedural activities in real world.
\newblock In \emph{CVPR}, 2024.

\bibitem[Kafle et~al.(2018)Kafle, Price, Cohen, and Kanan]{kafle2018dvqa}
Kushal Kafle, Brian Price, Scott Cohen, and Christopher Kanan.
\newblock Dvqa: Understanding data visualizations via question answering.
\newblock In \emph{CVPR}, 2018.

\bibitem[Kim et~al.(2021)Kim, Hong, Yim, Park, Yim, Hwang, Yun, Han, and Park]{kim2021donut}
Geewook Kim, Teakgyu Hong, Moonbin Yim, Jinyoung Park, Jinyeong Yim, Wonseok Hwang, Sangdoo Yun, Dongyoon Han, and Seunghyun Park.
\newblock Donut: Document understanding transformer without ocr.
\newblock \emph{arXiv}, 2021.

\bibitem[Lee et~al.(2012)Lee, Ghosh, and Grauman]{lee2012discovering}
Yong~Jae Lee, Joydeep Ghosh, and Kristen Grauman.
\newblock Discovering important people and objects for egocentric video summarization.
\newblock In \emph{CVPR}, 2012.

\bibitem[Lei et~al.(2018)Lei, Yu, Bansal, and Berg]{lei2018tvqa}
Jie Lei, Licheng Yu, Mohit Bansal, and Tamara~L Berg.
\newblock Tvqa: Localized, compositional video question answering.
\newblock In \emph{EMNLP}, 2018.

\bibitem[Li et~al.(2024{\natexlab{a}})Li, Zhang, Guo, Zhang, Li, Zhang, Zhang, Li, Liu, and Li]{li2024llavaov}
Bo~Li, Yuanhan Zhang, Dong Guo, Renrui Zhang, Feng Li, Hao Zhang, Kaichen Zhang, Yanwei Li, Ziwei Liu, and Chunyuan Li.
\newblock Llava-onevision: Easy visual task transfer.
\newblock \emph{arXiv}, 2024{\natexlab{a}}.

\bibitem[Li et~al.(2021)Li, Nagarajan, Xiong, and Grauman]{li2021ego}
Yanghao Li, Tushar Nagarajan, Bo~Xiong, and Kristen Grauman.
\newblock Ego-exo: Transferring visual representations from third-person to first-person videos.
\newblock In \emph{CVPR}, 2021.

\bibitem[Li et~al.(2024{\natexlab{b}})Li, Wang, and Jia]{li2024llamavid}
Yanwei Li, Chengyao Wang, and Jiaya Jia.
\newblock Llama-vid: An image is worth 2 tokens in large language models.
\newblock In \emph{ECCV}, 2024{\natexlab{b}}.

\bibitem[Lin et~al.(2024)Lin, Yin, Ping, Lu, Molchanov, Tao, Mao, Kautz, Shoeybi, and Han]{lin2023vila}
Ji~Lin, Hongxu Yin, Wei Ping, Yao Lu, Pavlo Molchanov, Andrew Tao, Huizi Mao, Jan Kautz, Mohammad Shoeybi, and Song Han.
\newblock Vila: On pre-training for visual language models.
\newblock In \emph{CVPR}, 2024.

\bibitem[Lin et~al.(2014)Lin, Maire, Belongie, Hays, Perona, Ramanan, Doll{\'a}r, and Zitnick]{lin2014COCO}
Tsung-Yi Lin, Michael Maire, Serge Belongie, James Hays, Pietro Perona, Deva Ramanan, Piotr Doll{\'a}r, and C~Lawrence Zitnick.
\newblock Microsoft coco: Common objects in context.
\newblock In \emph{ECCV}, 2014.

\bibitem[Liu et~al.(2023)Liu, Li, Wu, and Lee]{liu2023llava}
Haotian Liu, Chunyuan Li, Qingyang Wu, and Yong~Jae Lee.
\newblock Visual instruction tuning.
\newblock In \emph{NeurIPS}, 2023.

\bibitem[Lu et~al.(2022)Lu, Mishra, Xia, Qiu, Chang, Zhu, Tafjord, Clark, and Kalyan]{scienceqa}
Pan Lu, Swaroop Mishra, Tanglin Xia, Liang Qiu, Kai-Wei Chang, Song-Chun Zhu, Oyvind Tafjord, Peter Clark, and Ashwin Kalyan.
\newblock Learn to explain: Multimodal reasoning via thought chains for science question answering.
\newblock \emph{NeurIPS}, 2022.

\bibitem[Maaz et~al.(2024)Maaz, Rasheed, Khan, and Khan]{Maaz2023VideoChatGPT}
Muhammad Maaz, Hanoona Rasheed, Salman Khan, and Fahad~Shahbaz Khan.
\newblock Video-chatgpt: Towards detailed video understanding via large vision and language models.
\newblock In \emph{ACL}, 2024.

\bibitem[Mangalam et~al.(2023)Mangalam, Akshulakov, and Malik]{mangalam2023egoschema}
Karttikeya Mangalam, Raiymbek Akshulakov, and Jitendra Malik.
\newblock Egoschema: A diagnostic benchmark for very long-form video language understanding.
\newblock In \emph{NeurIPS Datasets and Benchmarks Track}, 2023.

\bibitem[Masry et~al.(2022)Masry, Long, Tan, Joty, and Hoque]{masry2022chartqa}
Ahmed Masry, Do~Xuan Long, Jia~Qing Tan, Shafiq Joty, and Enamul Hoque.
\newblock Chartqa: A benchmark for question answering about charts with visual and logical reasoning.
\newblock \emph{arXiv}, 2022.

\bibitem[Mathew et~al.(2021)Mathew, Karatzas, and Jawahar]{mathew2021docvqa}
Minesh Mathew, Dimosthenis Karatzas, and CV~Jawahar.
\newblock Docvqa: A dataset for vqa on document images.
\newblock In \emph{Proc. WACV}, 2021.

\bibitem[McKinzie et~al.(2024)McKinzie, Gan, Fauconnier, Dodge, Zhang, Dufter, Shah, Du, Peng, Weers, et~al.]{mckinzie2024mm1}
Brandon McKinzie, Zhe Gan, Jean-Philippe Fauconnier, Sam Dodge, Bowen Zhang, Philipp Dufter, Dhruti Shah, Xianzhi Du, Futang Peng, Floris Weers, et~al.
\newblock Mm1: Methods, analysis \& insights from multimodal llm pre-training.
\newblock \emph{arXiv}, 2024.

\bibitem[Merity et~al.(2016)Merity, Xiong, Bradbury, and Socher]{merity2016pointer}
Stephen Merity, Caiming Xiong, James Bradbury, and Richard Socher.
\newblock Pointer sentinel mixture models.
\newblock \emph{arXiv}, 2016.

\bibitem[OpenAI(2023)]{OpenAI_ChatGPT}
OpenAI.
\newblock Chat{GPT}: Optimizing language models for dialogue.
\newblock \url{https://openai.com/blog/chatgpt}, 2023.
\newblock Accessed: 2023.

\bibitem[Plizzari et~al.(2024)Plizzari, Goletto, Furnari, Bansal, Ragusa, Farinella, Damen, and Tommasi]{plizzari2024outlook}
Chiara Plizzari, Gabriele Goletto, Antonino Furnari, Siddhant Bansal, Francesco Ragusa, Giovanni~Maria Farinella, Dima Damen, and Tatiana Tommasi.
\newblock An outlook into the future of egocentric vision.
\newblock \emph{IJCV}, pp.\  1--57, 2024.

\bibitem[Shen et~al.(2023)Shen, Dudley, and Kristensson]{shen2023encode}
Junxiao Shen, John Dudley, and Per~Ola Kristensson.
\newblock Encode-store-retrieve: Enhancing memory augmentation through language-encoded egocentric perception.
\newblock \emph{arXiv}, 2023.

\bibitem[Sigurdsson et~al.(2018)Sigurdsson, Gupta, Schmid, Farhadi, and Alahari]{sigurdsson2018charadesego}
Gunnar~A. Sigurdsson, Abhinav Gupta, Cordelia Schmid, Ali Farhadi, and Karteek Alahari.
\newblock Charades-ego: A large-scale dataset of paired third and first person videos.
\newblock In \emph{arXiv}, 2018.

\bibitem[Song et~al.(2023)Song, Chai, Wang, Zhang, Zhou, Wu, Guo, Ye, Lu, Hwang, et~al.]{song2023moviechat}
Enxin Song, Wenhao Chai, Guanhong Wang, Yucheng Zhang, Haoyang Zhou, Feiyang Wu, Xun Guo, Tian Ye, Yan Lu, Jenq-Neng Hwang, et~al.
\newblock Moviechat: From dense token to sparse memory for long video understanding.
\newblock \emph{arXiv}, 2023.

\bibitem[Touvron et~al.(2023)Touvron, Lavril, Izacard, Martinet, Lachaux, Lacroix, Rozi{\`e}re, Goyal, Hambro, Azhar, et~al.]{touvron2023llama}
Hugo Touvron, Thibaut Lavril, Gautier Izacard, Xavier Martinet, Marie-Anne Lachaux, Timoth{\'e}e Lacroix, Baptiste Rozi{\`e}re, Naman Goyal, Eric Hambro, Faisal Azhar, et~al.
\newblock Llama: Open and efficient foundation language models.
\newblock \emph{arXiv}, 2023.

\bibitem[Vinyals et~al.(2015)Vinyals, Fortunato, and Jaitly]{vinyals2015pointer}
Oriol Vinyals, Meire Fortunato, and Navdeep Jaitly.
\newblock Pointer networks.
\newblock \emph{NeurIPS}, 28, 2015.

\bibitem[Wang et~al.(2024{\natexlab{a}})Wang, He, Hong, Cheng, Zhang, Qi, Gu, Huang, Xu, Dong, et~al.]{wang2024lvbench}
Weihan Wang, Zehai He, Wenyi Hong, Yean Cheng, Xiaohan Zhang, Ji~Qi, Xiaotao Gu, Shiyu Huang, Bin Xu, Yuxiao Dong, et~al.
\newblock Lvbench: An extreme long video understanding benchmark.
\newblock \emph{arXiv}, 2024{\natexlab{a}}.

\bibitem[Wang et~al.(2024{\natexlab{b}})Wang, Li, Li, Yu, He, Chen, Pei, Zheng, Xu, Wang, et~al.]{wang2024internvideo2}
Yi~Wang, Kunchang Li, Xinhao Li, Jiashuo Yu, Yinan He, Guo Chen, Baoqi Pei, Rongkun Zheng, Jilan Xu, Zun Wang, et~al.
\newblock Internvideo2: Scaling video foundation models for multimodal video understanding.
\newblock \emph{arXiv}, 2024{\natexlab{b}}.

\bibitem[Wu et~al.(2024)Wu, Li, Chen, and Li]{wu2024longvideobench}
Haoning Wu, Dongxu Li, Bei Chen, and Junnan Li.
\newblock Longvideobench: A benchmark for long-context interleaved video-language understanding.
\newblock \emph{arXiv}, 2024.

\bibitem[Xiao et~al.(2021)Xiao, Shang, Yao, and Chua]{xiao2021next}
Junbin Xiao, Xindi Shang, Angela Yao, and Tat-Seng Chua.
\newblock Next-qa: Next phase of question-answering to explaining temporal actions.
\newblock In \emph{CVPR}, 2021.

\bibitem[Xu et~al.(2023)Xu, Lan, Xie, Chen, and Lu]{xu2023retrieval}
Jiaqi Xu, Cuiling Lan, Wenxuan Xie, Xuejin Chen, and Yan Lu.
\newblock Retrieval-based video language model for efficient long video question answering.
\newblock \emph{arXiv}, 2023.

\bibitem[Xu et~al.(2024)Xu, Gao, Gan, Chen, Lai, Gang, Kang, and Dehghan]{xu2024slowfast}
Mingze Xu, Mingfei Gao, Zhe Gan, Hong-You Chen, Zhengfeng Lai, Haiming Gang, Kai Kang, and Afshin Dehghan.
\newblock Slowfast-llava: A strong training-free baseline for video large language models.
\newblock \emph{arXiv}, 2024.

\bibitem[Xue et~al.(2024)Xue, Chen, Li, Hu, Zhu, Li, Fang, Tang, Yang, Liu, et~al.]{xue2024longvila}
Fuzhao Xue, Yukang Chen, Dacheng Li, Qinghao Hu, Ligeng Zhu, Xiuyu Li, Yunhao Fang, Haotian Tang, Shang Yang, Zhijian Liu, et~al.
\newblock Longvila: Scaling long-context visual language models for long videos.
\newblock \emph{arXiv}, 2024.

\bibitem[Yang et~al.(2024)Yang, Yang, Hui, Zheng, Yu, Zhou, Li, Li, Liu, Huang, et~al.]{yang2024qwen2}
An~Yang, Baosong Yang, Binyuan Hui, Bo~Zheng, Bowen Yu, Chang Zhou, Chengpeng Li, Chengyuan Li, Dayiheng Liu, Fei Huang, et~al.
\newblock Qwen2 technical report.
\newblock \emph{arXiv}, 2024.

\bibitem[Yu et~al.(2019)Yu, Xu, Yu, Yu, Zhao, Zhuang, and Tao]{yu2019activitynet}
Zhou Yu, Dejing Xu, Jun Yu, Ting Yu, Zhou Zhao, Yueting Zhuang, and Dacheng Tao.
\newblock Activitynet-qa: A dataset for understanding complex web videos via question answering.
\newblock In \emph{AAAI}, 2019.

\bibitem[Zeki(2015)]{zeki2015area}
Semir Zeki.
\newblock Area v5—a microcosm of the visual brain.
\newblock \emph{Frontiers in integrative neuroscience}, 9:\penalty0 21, 2015.

\bibitem[Zhai et~al.(2023)Zhai, Mustafa, Kolesnikov, and Beyer]{zhai2023siglip}
Xiaohua Zhai, Basil Mustafa, Alexander Kolesnikov, and Lucas Beyer.
\newblock Sigmoid loss for language image pre-training.
\newblock In \emph{ICCV}, 2023.

\bibitem[Zhang et~al.(2023{\natexlab{a}})Zhang, Li, and Bing]{damonlpsg2023videollama}
Hang Zhang, Xin Li, and Lidong Bing.
\newblock Video-llama: An instruction-tuned audio-visual language model for video understanding.
\newblock \emph{arXiv}, 2023{\natexlab{a}}.
\newblock URL \url{https://arxiv.org/abs/2306.02858}.

\bibitem[Zhang et~al.(2023{\natexlab{b}})Zhang, Liu, Dong, Huang, Ling, Wang, Wang, and Qiao]{zhang2023movqa}
Hongjie Zhang, Yi~Liu, Lu~Dong, Yifei Huang, Zhen-Hua Ling, Yali Wang, Limin Wang, and Yu~Qiao.
\newblock Movqa: A benchmark of versatile question-answering for long-form movie understanding.
\newblock \emph{arXiv}, 2023{\natexlab{b}}.

\bibitem[Zhang et~al.(2024{\natexlab{a}})Zhang, Zhang, Li, Zeng, Yang, Zhang, Wang, Tan, Li, and Liu]{zhang2024longva}
Peiyuan Zhang, Kaichen Zhang, Bo~Li, Guangtao Zeng, Jingkang Yang, Yuanhan Zhang, Ziyue Wang, Haoran Tan, Chunyuan Li, and Ziwei Liu.
\newblock Long context transfer from language to vision.
\newblock \emph{arXiv}, 2024{\natexlab{a}}.

\bibitem[Zhang et~al.(2024{\natexlab{b}})Zhang, Li, Liu, Lee, Gui, Fu, Feng, Liu, and Li]{zhang2024llavanextvideo}
Yuanhan Zhang, Bo~Li, haotian Liu, Yong~jae Lee, Liangke Gui, Di~Fu, Jiashi Feng, Ziwei Liu, and Chunyuan Li.
\newblock Llava-next: A strong zero-shot video understanding model, April 2024{\natexlab{b}}.
\newblock URL \url{https://llava-vl.github.io/blog/2024-04-30-llava-next-video/}.

\bibitem[Zhu et~al.(2023)Zhu, Chen, Shen, Li, and Elhoseiny]{zhu2023minigpt}
Deyao Zhu, Jun Chen, Xiaoqian Shen, Xiang Li, and Mohamed Elhoseiny.
\newblock Minigpt-4: Enhancing vision-language understanding with advanced large language models.
\newblock \emph{arXiv}, 2023.

\end{thebibliography}
\bibliographystyle{iclr2025_conference}

\clearpage
\appendix

\section{More Analysis of Memory Pointer Prompting}
\begin{figure}[h]
    \centering
    \includegraphics[width=\textwidth,clip]
    {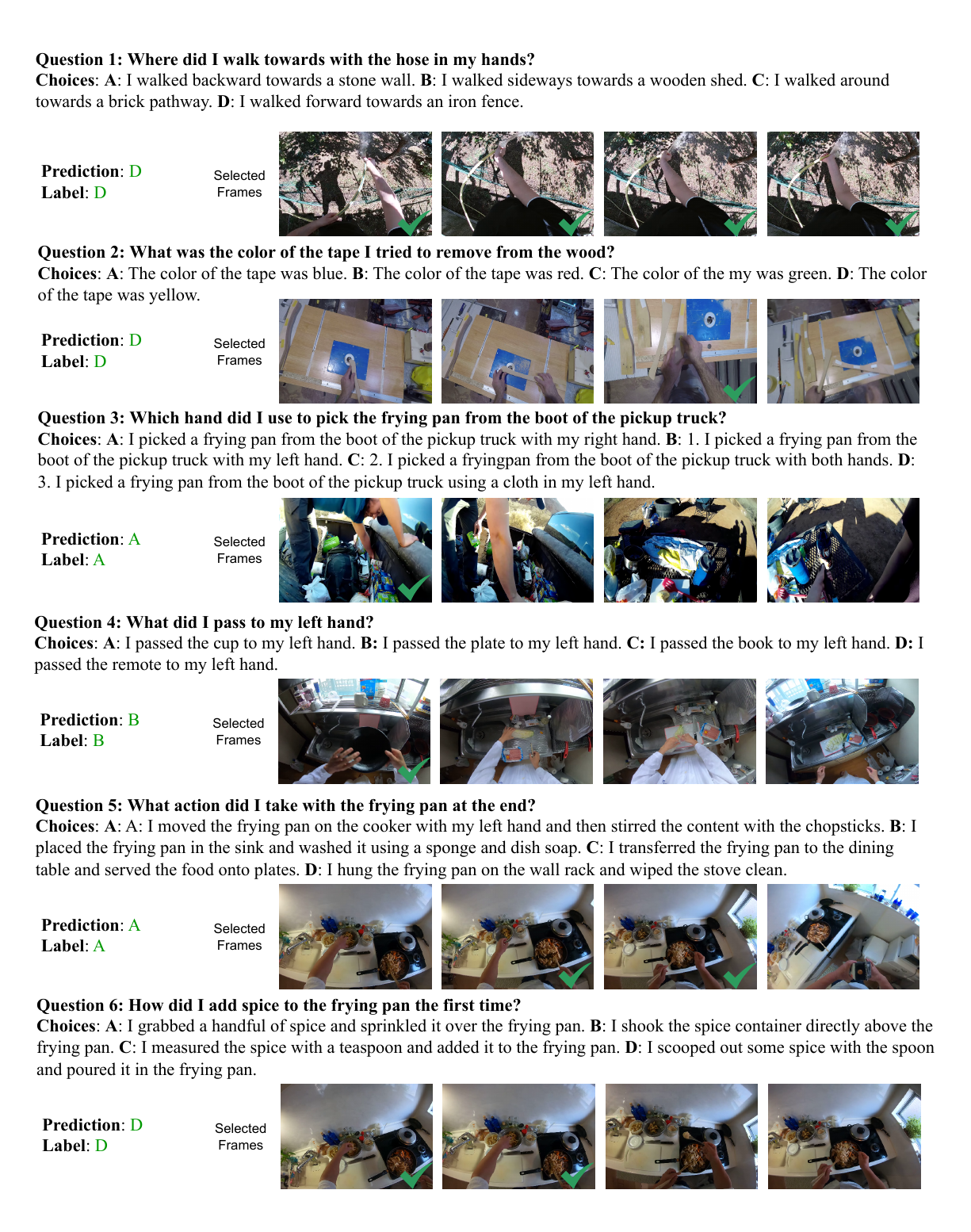}
    \vspace{-15pt}
    \caption{More key frame identification results of the global glimpse step on \benchname. We find high relevance between the identified key frames and the questions, demonstrating the effectiveness of the proposed Memory Pointer Prompting method.
    The \textcolor{green}{\checkmark} indicates that the selected frames are relevant to the questions.
    }
    \label{fig:frame_selection_more}
\end{figure}

To further assess the effectiveness of \modelname and the proposed Memory Pointer Prompting mechanism, we present additional visual results of key frame identification during the global glimpse step in Figure~\ref{fig:frame_selection_more}. 
\modelname demonstrates the capability to extract relevant visual information from a large set of frames based on the given questions.

\section{More Discussion on Related Works}
\textbf{Egocentric QA Data Generation}
After finishing the project, we find that the generation process in MM-Ego data engine shares some similar processes with the recently published LLaMA-VID~\citep{li2024llamavid} and GroundVQA~\citep{di2024groundvqa}. 
LLaMA-VID utilizes movie synopses to prompt LLMs to produce movie summaries and plot-related QA pairs.
GroundVQA generates short-term~(around 8 minutes) episodic memory QA from video narrations, but the goal and implementation details are different.
MM-Ego collects and processes videos with significantly more diverse video lengths from 30 seconds to 1 hour. 
The scale of our produced egocentric QA dataset is also significantly larger (7M vs. 303K).

\textbf{Long Video Understanding and Egocentric Understanding Evaluation Benchmarks}
In recent years, there have been some pioneering benchmarks for assessing the performance of multimodal models in understanding long videos~\citep{lei2018tvqa,song2023moviechat,zhang2023movqa,fu2024videomme,wang2024lvbench,wu2024longvideobench}.
Since the content of egocentric scenes differs from that of YouTube videos or movies, researchers have proposed specialized datasets for egocentric scene understanding.
There are benchmarks for egocentric images~\citep{cheng2024egothink} and videos~\citep{mangalam2023egoschema,di2024groundvqa,barmann2022qaego4d}. 
QAEgo4D~\citep{barmann2022qaego4d} benchmark assesses shorter-term video understanding (around 8 minutes) with a considerably smaller dataset (1,850 questions across 166 videos). 
EgoSchema~\citep{mangalam2023egoschema} has more video clips, yet the video lengths are still relatively short.
Concurrent with our work, HourVideo~\citep{chandrasegaran2024hourvideo} introduces an important egocentric QA benchmark consisting of 121,976 QA samples across 500 videos, ranging in length from 20 to 120 minutes.
Our EgoMemoria benchmark provides 7,026 QA samples for 629 videos and encompasses a wide range of video lengths, spanning from 30 seconds to 1 hour. The diverse video lengths make the benchmark significantly more challenging and closer to real-world egocentric video use cases.
Furthermore, we contribute a large-scale egocentric QA training dataset with more than 7 million QA samples for 8,933 egocentric videos, which enables further research on training more powerful egocentric video understanding models.
We show the statistics of these datasets in Table~\ref{tab:ego_benchmarks}.
Even when compared with other general/movie long video understanding benchmarks, the total numbers of QA samples and video counts in our Egomemoria benchmark are still significant.

\begin{table}[h!]
\centering
\tablestyle{4pt}{1.2}\scriptsize
\begin{tabular}{l|r|r|l|l}
\toprule
\textbf{Benchmark} & \textbf{Videos} & \textbf{QAs} & \textbf{Video Length Distribution} & \textbf{Data Type} \\
\shline
TVQA-test~\citep{lei2018tvqa} & 1,089 & 7,623 & 0 - 3 min & TV shows \\
MovieChat-1K-test~\citep{song2023moviechat} & 100 & 1,950 & 6 - 10 min & Movie \\ 
MoVQA~\citep{zhang2023movqa} & 20 & 21,953 & 7.5 - 120 min & Movie \\ 
Video-MME~\citep{fu2024videomme} & 900 & 2,700 & 0 - 60 min & Internet \\
LVBench~\citep{wang2024lvbench} & 103 & 1,549 & 30 - 140 min & Internet \\ 
LongVideoBench~\citep{wu2024longvideobench} & 3,763 & 6,678 & 0 - 60 min & Mixed \\ 
\hline
EgoSchema~\citep{mangalam2023egoschema} & 5,063 & 5,063 & 0.5 - 3 min & Egocentric \\ 
QAEgo4D-test~\citep{barmann2022qaego4d} & 166 & 1,850 & 0 - 8 min & Egocentric \\ 
GroundVQA-test (QAEgo4D\_close)~\citep{di2024groundvqa} & 148 & 500 & 0 - 8 min & Egocentric \\ 
HourVideo (Concurrent Work)~\citep{chandrasegaran2024hourvideo} & 500 & 12,976 & 20 - 120 min & Egocentric \\ 
\rowcolor{gray!15}
{MM-Ego Training} & 8,933 & 7M & 0.5 - 60 min & Egocentric \\ 
\rowcolor{gray!15}
{MM-Ego Evaluation (EgoMemoria Benchmark)} & 629 & 7,026 & 0.5 - 60 min & Egocentric \\ 
\bottomrule
\end{tabular}
\caption{Comparison with exiting long video understanding datasets.}
\label{tab:ego_benchmarks}
\end{table}

\noindent \textbf{Limitation and Future Work}
While \modelname demonstrates a strong ability in egocentric understanding, there is still room for further improvement. On the data and benchmark side, we can introduce more diverse egocentric understanding corpus~\citep{grauman2024ego,huang2024egoexolearn}. For the model itself, we plan to enhance its capacity to process a larger number of frames, such as at the order of thousands, to better handle longer or even always-on egocentric videos.

\section{Fine-tuning Exclusively on Egocentric QA Data}

To evaluate model performance when fine-tuning exclusively on egocentric QA data, we conduct an ablation study, with the results presented in Table~\ref{tab:ego_finetune_only}. For both LLaVA-OV~\citep{li2024llavaov} and MM-Ego, we consider two variants: one fine-tuned on our comprehensive data mixture and the other trained solely on egocentric QA data. The results demonstrate further performance improvements on the EgoMemoria benchmark. However, it is important to note that such domain-specific fine-tuning largely restricts the models' capacity for general video understanding.
On the other hand, we observe that MM-Ego still achieves significantly better performance in egocentric video understanding, attributed to the learning of the memory pointer prompting mechanism.

\begin{table}[h!]
\centering
\tablestyle{4pt}{1.2}\scriptsize
\begin{tabular}{l|c|c|c|c}
\toprule
\textbf{Name} & \textbf{Short} & \textbf{Medium} & \textbf{Long} & \textbf{Avg} \\ 
\shline
LLaVA-OV (Data Mixture) & 59.56 & 56.71 & 51.64 & 55.97 \\ 
\rowcolor{gray!15}
LLaVA-OV (Egocentric Data Only) & 62.58 & 58.13 & 53.61 & 58.11 \\ 
\hline
MM-Ego (Data Mixture) & 63.16 & 62.04 & 58.62 & 61.27 \\ 
\rowcolor{gray!15}
MM-Ego (Egocentric Data Only) & {65.41} & {64.89} & {61.05} & {63.78} \\ 
\bottomrule
\end{tabular}
\caption{Performance comparison of different models on EgoMemoria (MDA) with fine-tuning exclusively on egocentric QA data.}
\label{tab:ego_finetune_only}
\end{table}

\section{Analysis of Using Language Model in Data Engine}

The motivation for using a language model to convert egocentric video captions into egocentric QA conversations is to address the ``chicken-or-egg dilemma''. If we rely on a vision-language model~(VLM) to generate egocentric QA pairs, the quality of the data is inherently limited by the egocentric understanding capabilities of the labeling VLM. Consequently, downstream VLMs trained on this synthetic data cannot outperform the labeling VLM. This creates the chicken-or-egg problem: do we have a strong egocentric VLM first, or do we have good egocentric QA data first?

To address this dilemma, our ``narration-to-egocentric QA'' data engine leverages a language-only model to generate QA samples. This approach circumvents the hallucination and inaccuracy issues often associated with long video understanding, which is a very challenging task by itself. Since we have access to high-quality, fine-grained, and densely annotated video narrations created by human labelers in Ego4D, the essential visual information has already been effectively translated into the narrations.

We conduct a preliminary experiment of using image frames augmented GPT-4o to produce egocentric QA. We uniformly sample 32 frames with a resolution of 336x448 from a 30 seconds egocentric video.

\textbf{Input Narrations:}
VID\_NARRATION = [``I close the tap.'', ``I press the buttons on the rice cooker.'', ``I open the rice cooker.'', ``I stir the rice.'', ``I close the rice cooker.'', ``I press the buttons on the rice cooker.'']

\textbf{Text Prompts:}
\texttt{
``Please design at most 5 questions and answers about self-questioning my memory based on a video. For example, ask questions about what I have done, what objects did I interact, where did I go, what did other people do, the locations, colors, and shapes of the objects in the scenes if mentioned, based solely on the video narrations \$VID\_NARRATION. The answers should be in first-person perspective or objective description of the objects. Remember this rule: special word 'I', 'c', or 'C' represent me (first-person perspective), special word 'x', 'y', 'Y', 'Z', 'z', 'o', 'O', 'X', etc. represent somebody else, this is very important. Ask very diverse questions that you are sure about the answers, and the questions cannot be answered without watching the video, which means no common sense questions. Don't generate anything if you do not have enough information. Attach index of the sentence that answer is based upon. Directly reply by strictly following this example: 'QUESTION: Where did I leave my key? ANSWER:I put it on the table.\_0,2 QUESTION:Did I close the door when I left? ANSWER:Yes I did.\_3 QUESTION:What is the color of the cup? ANSWER:White.\_1''
}

\textbf{Generated Egocentric QAs:}
We present the generated QAs by GPT-4o both with and without video input in Table~\ref{tab:questions_answers_2} and Table~\ref{tab:questions_answers}.
In this example, we observe that the generated QAs have similar quality whether videos are included or not. However, incorporating video into GPT-4o's input significantly increases the time and computational costs. Moreover, when applying VLM to a large volume of videos, it becomes challenging to verify if any visual hallucinations arise from the video understanding process. Therefore, we choose to use a mature pure language model in our data engine.

\begin{table}[h!]
\centering
\renewcommand{\arraystretch}{1.5} %
\setlength{\tabcolsep}{8pt} %
\begin{tabular}{|p{0.5\textwidth}|p{0.4\textwidth}|}
\hline
\textbf{Question} & \textbf{Answer} \\ \hline
What did I do to the tap in the video? & I closed the tap. \\ \hline
How did I interact with the rice cooker buttons? & I pressed the buttons on the rice cooker twice. \\ \hline
What did I use to stir the rice? & I used a spoon to stir the rice. \\ \hline
Did I leave the rice cooker open after stirring the rice? & No, I closed the rice cooker. \\ \hline
At what points in the video was the rice cooker interacted with? & The rice cooker was opened, stirred, and closed. \\ \hline
\end{tabular}
\caption{GPT-4o generated egocentric QAs with video input.}
\label{tab:questions_answers_2}
\end{table}

\begin{table}[h!]
\centering
\renewcommand{\arraystretch}{1.5} %
\setlength{\tabcolsep}{8pt} %
\begin{tabular}{|p{0.6\textwidth}|p{0.3\textwidth}|}
\hline
\textbf{Question} & \textbf{Answer} \\ \hline
What action did I perform on the tap? & I closed it. \\ \hline
What did I do after pressing the buttons on the rice cooker the first time? & I opened the rice cooker. \\ \hline
What did I use to stir the rice? & I stirred the rice. \\ \hline
How many times did I press the buttons on the rice cooker? & Twice. \\ \hline
What sequence of actions involved the rice cooker? & I pressed the buttons, opened it, stirred the rice, closed it, and pressed the buttons again. \\ \hline
\end{tabular}
\caption{Generated egocentric QAs without video input.}
\label{tab:questions_answers}
\end{table}

\section{MM-Ego's Connection with Cognitive Neuroscience}
Our proposed ``Memory Pointer Prompting'' two-step mechanism is inspired by the way human naturally process videos. When answering a specific question about a long video, we typically start by quickly scanning the entire video to identify frames relevant to the question, which is similar to our “global glimpse” step. We then closely examine those frames to find the answers, which resembles our “fallback” step.

In the study of human visual perception system, researchers identify two distinct pathways for processing visual information in the brain: the magnocellular and parvocellular pathways~\citep{zeki2015area,hochstein2002view}.
Our ``global glimpse" step mirrors the functionality of the magnocellular pathway which is responsible for handling information about large, fast-moving objects. On the other hand, our ``fallback" step aligns with the role of the parvocellular pathway which specializes in processing details of small, slow-moving objects.

\section{Future Direction on Processing Longer Videos}
To further enhance MM-Ego's capability to handle even longer videos, we can adopt two strategies. First, leveraging aggressive parallelism techniques, such as sequential and tensor parallelism~\citep{xue2024longvila}, can significantly extend the context length of the transformer model. This will extend the model’s ability to do reasoning in more frames. Second, we can introduce a hierarchical structure to the compressed visual embeddings by further consolidating embeddings from multiple frames into a single representation. Then we can design multiple global glimpse steps, enabling the model to identify relevant frames in a coarse-to-fine manner.

\end{document}